\documentclass[]{template}

\usepackage{makecell}
\usepackage[normalem]{ulem}
\usepackage{nicefrac}
\usepackage{float}
\usepackage{placeins}
\usepackage{multirow}
\usepackage{array}
\usepackage{listings}

\setboolean{logo}{true}
\setlist{nosep,leftmargin=1.25em}
\renewcommand{\arraystretch}{0.95}
\hypersetup{
  pdftitle={SciDocBench: A Workflow-Centered Benchmark and Data Pipeline for Scientific Document Understanding},
  pdfauthor={Shenxi Wu, Yuhong Liu, Haosong Zhang, Tongjin Zou, Yanxun Zhang, Gaochang Chen, Liang Dun, Jiaqi Wang, Zhecan James Wang, Yuhang Zang, Dahua Lin}
}
\renewcommand{\absfont}{\linespread{1.05}\fontsize{9.5}{11}\selectfont}

\title{SciDocBench: A Workflow-Centered Benchmark and Data Pipeline for Scientific Document Understanding}

\newcommand{\equalcontrib}{\textsuperscript{*}}
\newcommand{\corresponding}{\textsuperscript{\textdagger}}
\newcommand{\authorfootnotes}{%
  \begingroup
  \renewcommand{\thefootnote}{\fnsymbol{footnote}}%
  \footnotetext[1]{Equal contribution.}%
  \footnotetext[2]{Corresponding author: Dahua Lin
  (\href{mailto:dhlin@ie.cuhk.edu.hk}{dhlin@ie.cuhk.edu.hk}).}%
  \endgroup
}
\author[1,2]{Shenxi Wu\equalcontrib}
\author[1,2]{Yuhong Liu\equalcontrib}
\author[3]{Haosong Zhang}
\author[4]{Tongjin Zou}
\author[4]{Yanxun Zhang}
\author[5]{Gaochang Chen}
\author[6]{Liang Dun}
\author[7]{Jiaqi Wang}
\author[2]{Zhecan James Wang}
\author[2]{Yuhang Zang}
\author[1,2,8]{Dahua Lin\corresponding}
\affil[1]{The Chinese University of Hong Kong}
\affil[2]{Shanghai Artificial Intelligence Laboratory}
\affil[3]{New York University}
\affil[4]{Fudan University}
\affil[5]{Shanghai Jiao Tong University}
\affil[6]{Harbin Institute of Technology}
\affil[7]{JD Explore Academy}
\affil[8]{Centre for Perceptual and Interactive Intelligence (CPII) Limited}

\newcommand{\ours}{\textsc{SciDocBench}}
\newcommand{\dataset}{\textsc{SciDocDataset}}
\newcommand{\ir}{\textsc{SciDocIR}}

\newcolumntype{Y}{>{\raggedright\arraybackslash}X}
\newcolumntype{C}{>{\centering\arraybackslash}X}
\definecolor{SciGreen}{RGB}{90,150,70}
\definecolor{SciGreenLight}{RGB}{235,247,232}
\definecolor{SciBlueLight}{RGB}{235,244,255}
\definecolor{SciOrangeLight}{RGB}{255,245,232}
\definecolor{SciGray}{RGB}{245,245,245}
\definecolor{scidocgreen}{RGB}{234,247,232}
\definecolor{scidocblue}{RGB}{233,243,255}
\definecolor{scidocorange}{RGB}{255,244,229}
\definecolor{scidarkorange}{RGB}{255,190,1}
\definecolor{scilightorange}{RGB}{255,228,171}
\definecolor{sciweakorange}{RGB}{250,235,215}
\definecolor{scidarkgreen}{RGB}{50,150,0}
\definecolor{scilightred}{RGB}{250,128,114}
\definecolor{scicheckgreen}{HTML}{0B8F5A}
\definecolor{scicrossred}{HTML}{C43D3D}
\definecolor{scipartialamber}{HTML}{B07A00}

\definecolor{sciRankI}{HTML}{F7D9A0}
\definecolor{sciRankII}{HTML}{FBEACB}
\definecolor{sciRankIII}{HTML}{FDF6E9}
\newcommand{\rI}[1]{\cellcolor{sciRankI}\textbf{#1}}
\newcommand{\rII}[1]{\cellcolor{sciRankII}#1}
\newcommand{\rIII}[1]{\cellcolor{sciRankIII}#1}
\newcommand{\mlogo}[1]{\makebox[1.35em][l]{%
  \raisebox{-0.22\height}{\includegraphics[height=1em]{figures/model_logos/#1}}}}
\newcommand{\nologo}{\makebox[1.35em][l]{}}
\newcommand{\hd}[1]{\makecell[bc]{#1}}
\newcommand{\gain}[1]{\textcolor{scicheckgreen}{#1}}
\newcommand{\loss}[1]{\textcolor{scicrossred}{#1}}

\newcommand{\yesmark}{\textcolor{scicheckgreen}{\ding{51}}}
\newcommand{\nomark}{\textcolor{scicrossred}{\ding{55}}}
\newcommand{\scipartmark}{\textcolor{scipartialamber}{\textbf{--}}}

\lstdefinestyle{scidocir}{
    basicstyle=\ttfamily\footnotesize,
    breaklines=true,
    frame=single,
    columns=fullflexible,
    keepspaces=true,
    showstringspaces=false,
}
\lstdefinestyle{scidocseed}{
    basicstyle=\ttfamily\scriptsize,
    breaklines=true,
    breakatwhitespace=true,
    frame=single,
    framerule=0.35pt,
    rulecolor=\color{black!35},
    backgroundcolor=\color{SciGray},
    columns=fullflexible,
    keepspaces=true,
    showstringspaces=false,
    xleftmargin=0.8em,
    xrightmargin=0.8em,
    aboveskip=0.45em,
    belowskip=0.65em,
}

\begin{abstract}
Scientific papers require models to integrate evidence across text, equations, figures, tables, code, and datasets while preserving its provenance. Beyond answer correctness, scientific reading requires verifiable outputs from operations such as evidence localization, definition extraction, and consistency checking. We introduce \textsc{SciDocBench}, a workflow-centered benchmark targeting these operations through 124 expert-authored and difficulty-screened questions across seven research-assistant capability groups, 19 subtasks, and five scientific domains. Each question is instantiated in four matched settings formed by pairing its bilingual variants with the All Images First and Markdown Interleaved document representations, yielding 496 evaluation instances. The strongest evaluated model, Claude-Opus-5, scores 62.6 out of 100, with remaining gaps in evidence localization, structured information extraction, cross-document synthesis, and robustness to document representation. To convert these diagnostics into scalable training signals, we introduce \textsc{SciDocIR}, a structured representation of scientific document objects, layout and cross-reference relations, and provenance. Using \textsc{SciDocIR}, we construct \textsc{SciDocDataset}, which contains 4K supervised fine-tuning instances and 10K reinforcement-learning instances, built on 14 verifiable training subtasks.
Post-training Qwen3.6-27B on task-aligned data improves its \textsc{SciDocBench} score from 40.03 to 45.33 with supervised fine-tuning and to 45.74 with subsequent reinforcement learning. Both adapted models preserve DocVQA and InfoVQA performance and improve ChartQA accuracy over the original model by 0.80 and 3.40 points, respectively. Together, \textsc{SciDocBench}, \textsc{SciDocIR}, and \textsc{SciDocDataset} connect capability diagnosis with verifiable training-data construction for scientific-document assistants.%
\par\smallskip\noindent
\textbf{Code and data:}
\href{https://github.com/InternLM/SciDocBench}{GitHub}
\enspace\textbar\enspace
\href{https://huggingface.co/datasets/HenryExcellent/SciDocBench}{Benchmark}
\enspace\textbar\enspace
\href{https://huggingface.co/datasets/HenryExcellent/SciDocBench-Training-Data}{Training data}.
\end{abstract}

\begin{document}
\raggedbottom

\maketitle
\authorfootnotes

\begin{figure}[!tp]
    \centering
    \includegraphics[width=1.0\linewidth]{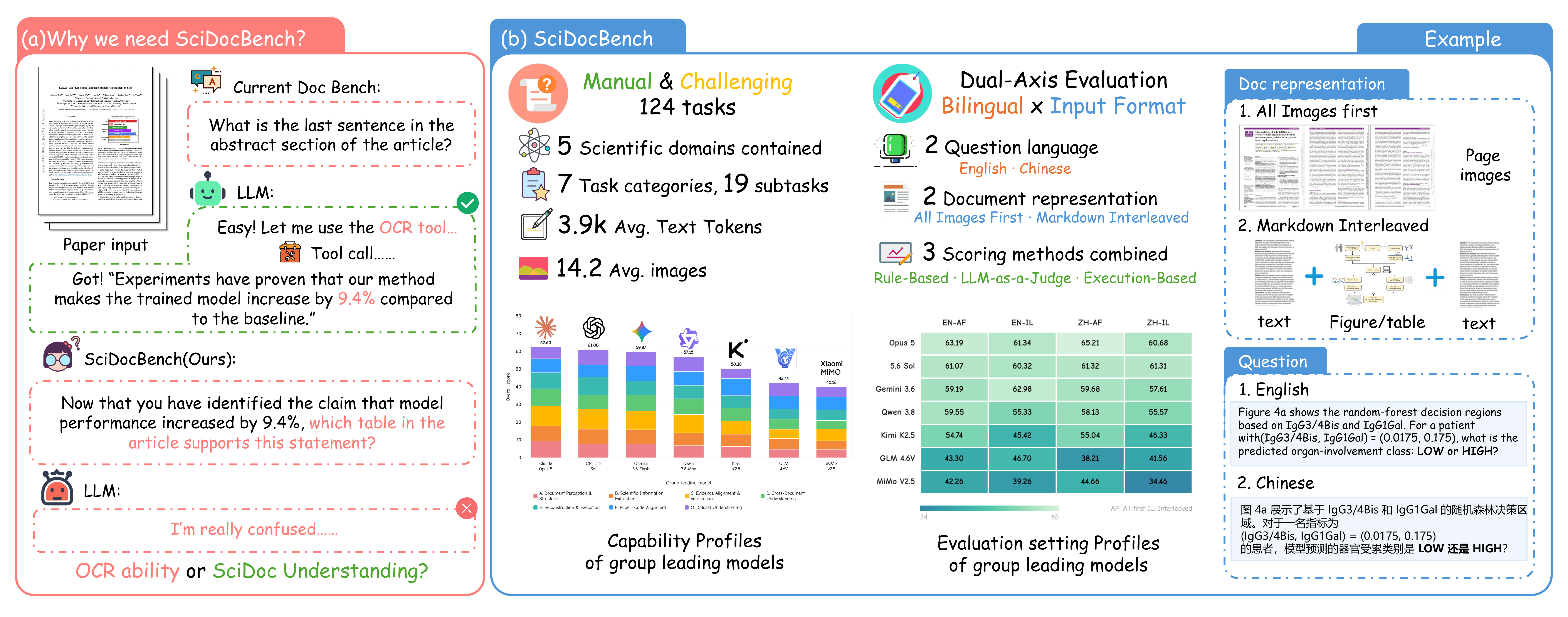}
    \captionsetup{font=footnotesize,labelfont=bf,skip=3pt}
    \caption{\textbf{Overview of \ours{}}.
    \textbf{(a) From text extraction to evidence grounding.} A direct text-extraction prompt can be answered by recovering visible content, whereas the \ours{} example additionally requires locating the document evidence supporting a scientific claim.
    \textbf{(b) Benchmark and evaluation design.} \ours{} contains \textbf{124 expert-authored and difficulty-screened questions} across five scientific domains, seven capability groups, and 19 subtasks. Its \textbf{dual-axis protocol} pairs bilingual question variants with the \textit{All Images First} and \textit{Markdown Interleaved} document representations, producing \textbf{four matched settings} per question. Responses are assessed using task-appropriate rule-based, LLM-as-a-judge, or execution-based evaluators. The plots summarize capability-group and evaluation-setting profiles for group-leading models, while the example illustrates the matched question and document-representation variants.}
    \label{fig:teaser}
\end{figure}

\section{Introduction}

Scientific papers communicate claims through text, equations, figures, tables, code, and datasets~\citep{xia2024docgenome,li2024m3sciqa}. A useful scientific assistant must do more than \textit{retrieve passages or summarize prose}: it must locate evidence, recover experimental details, verify numerical relations, compare papers, trace provenance, and produce reusable outputs~\citep{dasigi2021qasper,starace2025paperbench}. These operations underpin literature review, claim verification, and reproducibility analysis.

Scientific document understanding presents three challenges. \textbf{Heterogeneous structure:} scientific objects depend on layout, captions, reading order, and cross-references~\citep{ma2024mmlongbenchdoc}. \textbf{Compositional reasoning:} verifying one claim can require evidence from several objects, papers, or repositories~\citep{li2024m3sciqa}. \textbf{Verifiable outputs:} answers must preserve numerical fidelity, structure, and provenance, while evaluation must distinguish task competence from sensitivity to language and document representation.

Existing benchmarks cover document QA, layout analysis, scientific QA, and chart reasoning~\citep{mathew2021docvqa,xu2020layoutlm,dasigi2021qasper,masry2022chartqa}. Many emphasize whether a model returns the correct answer; others also test supporting evidence or individual document operations. A complementary question is whether a model can produce the \textbf{verifiable intermediate deliverables} needed in scientific reading: an evidence location, a definition with its context, a consistency check, or an integrated record with provenance (Table~\ref{tab:benchmark_comparison_focus}). These deliverables expose failures that answer correctness alone may leave unresolved.

Figure~\ref{fig:teaser} illustrates the distinction: identifying a claim is only part of the task when answering it also requires locating and interpreting the supporting table. We introduce \ours{}, a \textbf{workflow-centered benchmark} with 124 expert-authored, difficulty-screened questions across five scientific domains, seven capability groups, and 19 subtasks. Its tasks bring evidence-handling operations into the foreground: models must locate support, interpret results under the relevant definitions, and reconcile information across papers. These operations become evaluation targets through question-specific deliverables, such as evidence attribution, structured extraction, or cross-paper synthesis.

Our \textbf{matched dual-axis protocol} pairs English and Chinese questions with \textit{All Images First} or \textit{Markdown Interleaved} documents, producing 496 instances with four settings. Task semantics, reference answers, and evaluation criteria remain fixed, isolating language and representation effects. Structured or executable outputs are assessed using rules, semantic judgment, or execution.

To turn diagnosis into training supervision, we introduce \ir{}, a normalized representation that aligns visual and semantic document objects while preserving layout, cross-references, and provenance. It supports \dataset{}, which contains approximately 4K SFT and 10K RL instances, built on \textbf{14 verifiable training subtasks}. Targets are checked through source records, controlled perturbations, exact operations, or execution. Benchmark questions are not converted into training examples.

Across 13 proprietary and open models, Claude-Opus-5 achieves the highest overall score of \textbf{62.6 out of 100}; no model leads more than two capability groups. Scientific extraction and cross-document synthesis remain difficult. For 10 models, the representation gap exceeds the language gap, showing that equivalent evidence can elicit different outcomes depending on its presentation.

We further examine whether task-aligned data improves a scientific-document assistant. Using Qwen3.6-27B, SFT and SFT followed by GRPO raise the \ours{} score from 40.03 to 45.33 and 45.74, gains of 5.30 and 5.71 points over the original model. Both adapted models preserve general document understanding on DocVQA and InfoVQA and improve ChartQA by 0.80 and 3.40 points, respectively. These experiments demonstrate the training utility of the data pipeline alongside its role in benchmark diagnosis.

Our contributions are as follows: \textbf{1) Workflow-centered benchmark.} We introduce \ours{}, a workflow-centered benchmark that evaluates concrete scientific research tasks across full documents, multiple papers, and associated code or data, using a matched bilingual and dual-representation protocol. \textbf{2) Systematic model evaluation.} We evaluate proprietary and open multimodal models, revealing complementary capability profiles, persistent weaknesses in structured extraction and cross-document synthesis, and substantial sensitivity to document representation. \textbf{3) Evaluation-to-training resources.} We construct \ir{} and \dataset{}, connecting benchmark diagnostics with scalable, verifiable training data. SFT and SFT followed by GRPO improve scientific-document performance while preserving general document question answering.

\section{Related Work}

\noindent \textbf{Document understanding benchmarks.}
PubLayNet and LayoutLM support layout annotation and layout-aware representations~\citep{zhong2019publaynet,xu2020layoutlm}; DocVQA and DUE evaluate document-image QA~\citep{mathew2021docvqa,borchmann2021due}. MMLongBench-Doc extends evaluation to long multimodal documents~\citep{ma2024mmlongbenchdoc}, and ArXivDoc compares image, text, and interleaved scientific-document representations~\citep{khalighinejad2026document}. \ours{} examines how these perceptual and reading capabilities compose within scientific workflows.

\noindent \textbf{Scientific question answering and research assistance.}
QASPER provides full-paper questions and supporting evidence~\citep{dasigi2021qasper}, while PubMedQA focuses on biomedical abstracts~\citep{jin2019pubmedqa}. M3SciQA and PaperScope address multimodal, multi-document QA and deep research~\citep{li2024m3sciqa,xiong2026paperscope}; PaperBench evaluates end-to-end research replication~\citep{starace2025paperbench}. \ours{} complements these resources with explicit tests of claim grounding, notation extraction, consistency checking, and structured result integration.

\noindent \textbf{Chart, table, and scientific-object reasoning.}
PlotQA and ChartQA evaluate graphical and numerical reasoning~\citep{methani2020plotqa,masry2022chartqa}; SCITAB tests compositional verification over scientific tables~\citep{lu2023scitab}. \ours{} places such operations within full documents, where evidence may span a figure, caption, equation, appendix, or another paper and must retain its provenance.

\noindent \textbf{Structured scientific-document resources and multimodal training data.}
S2ORC and DocGenome structure scholarly text, references, and multimodal document objects~\citep{lo2020s2orc,xia2024docgenome}. Visual instruction tuning and MM-IFEngine use generated multimodal supervision and output constraints~\citep{liu2023llava,ding2025mmifengine}, while DocSeeker combines SFT with evidence-aware RL~\citep{yan2026docseeker}. Our framework emphasizes task-specific verifiability: \ir{} aligns rendered PDF content with available LaTeX semantics, and \dataset{} uses source records and controlled operations to construct training targets.

\begin{table}[!tp]
\centering
\caption{
Focused comparison with representative scientific-document benchmarks.
\yesmark denotes primary coverage, \scipartmark partial or adjacent
coverage, and \nomark that the dimension is not an evaluation target.
\textbf{Cross-doc.} requires integrating multiple documents;
\textbf{code/data prov.} denotes explicit reasoning over code,
datasets, or provenance.
}
\label{tab:benchmark_comparison_focus}

\scriptsize
\setlength{\tabcolsep}{3pt}
\renewcommand{\arraystretch}{1.06}

\begin{tabularx}{\linewidth}{@{}l Y c c c c@{}}
\toprule
\textbf{Benchmark}
& \textbf{Primary focus}
& \makecell{\textbf{Full paper}\\\textbf{or PDF}}
& \makecell{\textbf{Scientific}\\\textbf{objects}}
& \makecell{\textbf{Cross-}\\\textbf{doc.}}
& \makecell{\textbf{Code/data}\\\textbf{prov.}} \\
\midrule

DocVQA~\citeyearpar{mathew2021docvqa}
& Document-image question answering
& \nomark & \nomark & \nomark & \nomark \\

QASPER~\citeyearpar{dasigi2021qasper}
& Evidence-grounded full-paper QA
& \yesmark & \scipartmark & \nomark & \nomark \\

ChartQA~\citeyearpar{masry2022chartqa}
& Chart question answering
& \nomark & \scipartmark & \nomark & \nomark \\

SCITAB~\citeyearpar{lu2023scitab}
& Scientific-table claim verification
& \nomark & \yesmark & \nomark & \nomark \\

MMLongBench-Doc~\citeyearpar{ma2024mmlongbenchdoc}
& Long-context document multimodal QA
& \yesmark & \scipartmark & \nomark & \nomark \\

DocGenome~\citeyearpar{xia2024docgenome}
& Scientific-document structure and QA
& \yesmark & \yesmark & \nomark & \nomark \\

M3SciQA~\citeyearpar{li2024m3sciqa}
& Multimodal multi-document QA
& \yesmark & \yesmark & \yesmark & \nomark \\

PaperBench~\citeyearpar{starace2025paperbench}
& End-to-end research replication
& \yesmark & \scipartmark & \nomark & \yesmark \\

ArXivDoc~\citeyearpar{khalighinejad2026document}
& Scientific retrieval across representations
& \yesmark & \yesmark & \nomark & \nomark \\

PaperScope~\citeyearpar{xiong2026paperscope}
& Agentic multi-document deep research
& \yesmark & \yesmark & \yesmark & \nomark \\

\midrule
\rowcolor{scidocgreen}
\textbf{\ours}
& \textbf{Workflow-centered scientific-document understanding}
& \yesmark & \yesmark & \yesmark & \yesmark \\

\bottomrule
\end{tabularx}
\end{table}

\noindent \textbf{Positioning and scope.}
Table~\ref{tab:benchmark_comparison_focus} compares full-document input, scientific-object reasoning, cross-document understanding, and code or data provenance. Partial coverage denotes a related signal rather than a central evaluation target. \ours{} organizes selected scientific-reading operations around \textbf{evidence requirements and verifiable deliverables}, using matched inputs to diagnose where these operations break down.

\section{The SciDocBench Benchmark}
\label{sec:benchmark}

\subsection{Benchmark Overview and Design Principles}

\ours{} contains 124 expert-authored questions in seven capability groups and 19 subtasks. English and Chinese variants are paired with two document representations, yielding 496 matched instances that test evidence localization, scientific-object interpretation, verification, cross-document integration, and connections to code and data.

Three principles guide construction. \textbf{Workflow realism:} questions derive from concrete activities such as reviewing claims, integrating results, or reconstructing figures. \textbf{Evidence fidelity and controlled variation:} matched settings preserve task-relevant content while varying language and representation. \textbf{Verifiable structure:} questions request structured or executable deliverables when the research activity demands them, enabling task-appropriate evaluation.

\subsection{Task Taxonomy}
\label{sec:taxonomy}

\ours{} organizes its questions into seven capability groups. \textbf{Document Perception and Structure}, \textbf{Scientific Information Extraction}, and \textbf{Evidence Alignment and Verification} evaluate whether models can locate, interpret, extract, and check information within scientific documents. \textbf{Cross-Document Understanding} evaluates the integration of notation, datasets, and results across papers. \textbf{Reconstruction and Execution}, \textbf{Paper--Code Alignment}, and \textbf{Dataset Understanding} examine whether models can transform scientific content into reusable artifacts and connect papers with associated computational resources. The complete inventory of 19 subtasks, together with their definitions and evaluation focus, is provided in Appendix~\ref{app:tasks}.

\subsection{Evaluation Instances and Scoring}
\label{sec:benchmark_contract}

\noindent \textbf{Model inputs.}
Each instance supplies all task-relevant source materials followed by an English or Chinese question. All Images First presents every paper page as an image; Markdown Interleaved arranges extracted text and visual elements in document order. Multi-document tasks supply all required papers. Paper--code and dataset tasks may additionally include file trees, code excerpts, or dataset descriptions. Reference answers, task labels, and evaluation metadata are withheld.

\noindent \textbf{Output contract.}
The question specifies the requested deliverable and any exact format requirements. Outputs include labels, values, spans, ordered lists, JSON, tables, graphs, reconstructed data, LaTeX, or programs. These formats reflect scientific outputs that must be parsed, verified, or reused, rather than arbitrary formatting constraints.

\noindent \textbf{Evaluation and score aggregation.}
We assign each underlying question one of three evaluators: deterministic rules for 57 questions, a semantic LLM judge for 66, and execution for one image-transformation task. Rules compare specified fields, values, or orderings; the judge applies the question-specific rubric to the reference and final answer; execution compares generated and reference outputs. Appendix~\ref{apx:scoring_protocol} defines partial credit and failure handling and lists the evaluator for every question. Each instance receives a score between zero and one; failed, empty, or unusable responses receive zero. The overall score averages all 496 instances, scaled to 0--100.

Capability-group and setting scores average their respective instances on the same scale. All four variants preserve source content, task and reference-answer semantics, and scoring criteria, enabling paired sensitivity analysis. Scores assess final answers, excluding internal reasoning traces; an explanation explicitly requested in the final answer remains part of its output contract.

\subsection{Benchmark Construction}
\label{sec:benchmark_construction}

\begin{figure}[!tp]
    \centering
    \includegraphics[width=1.0\linewidth]{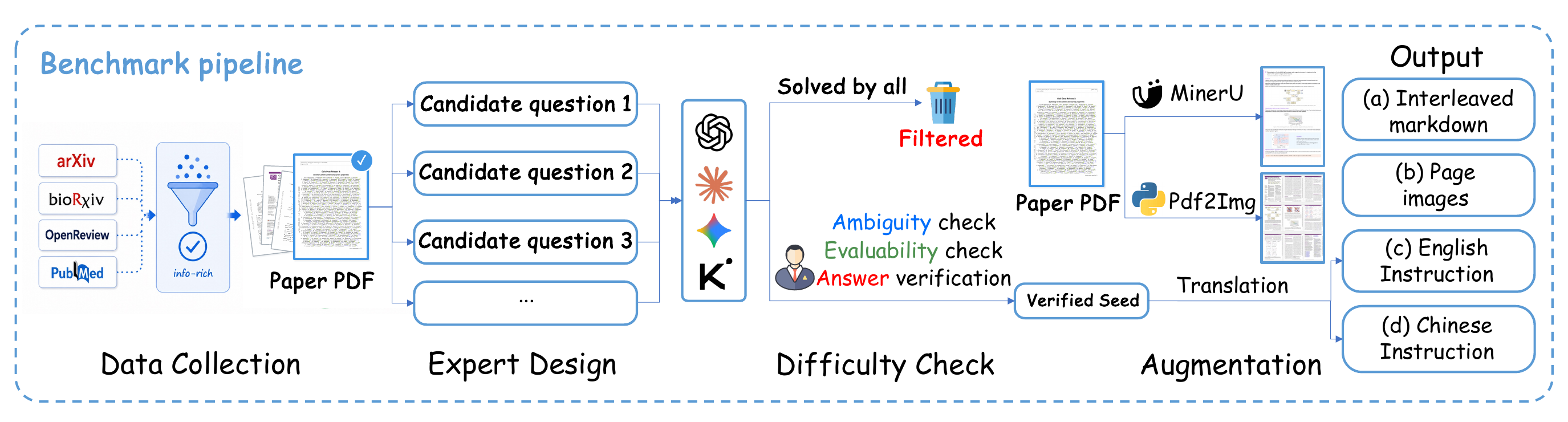}
    \caption{\textbf{\ours{} construction.} Information-rich papers support expert-authored research tasks. Difficulty screening and independent verification reject trivial or ambiguous items. Bilingual and dual-representation augmentation produces four matched instances per accepted question.}
    \label{fig:benchmark_pipeline}
\end{figure}

Figure~\ref{fig:benchmark_pipeline} summarizes paper selection, expert authoring and review, difficulty screening, and controlled augmentation. Questions must be scientifically meaningful, answerable from the supplied materials, unambiguous, and challenging. Appendix~\ref{examples} provides representative examples.

\noindent \textbf{Paper selection.}
We collect 116 publicly accessible papers spanning five scientific domains from arXiv, bioRxiv, OpenReview, PubMed, and corresponding journal pages. Selection favors information-rich papers with figures, tables, equations, captions, appendices, and experimental descriptions; we record these object types and counts to characterize document richness.

Papers supporting only generic summarization or unstable answers are excluded. Theoretical papers support notation and equation tasks; empirical papers support experimental-setting extraction, table consistency, and result integration. Appendix~\ref{app:tasks} records source papers and construction statistics.

\noindent \textbf{Question authoring and review.}
Each candidate records its source documents, question, reference answer, task labels, evidence locator, and evaluator. A reviewer other than the author checks answerability, reference correctness, and instruction and output clarity. Structured-output requirements appear in the question itself. Appendix~\ref{apx:annotation_verification} documents field-level requirements and the independent-review interfaces.

\subsection{Quality Control and Evaluation Reliability}

\paragraph{Difficulty screening.}
To remove questions already solved by strong systems, we evaluate each candidate once using Claude Opus 4.8~\citep{anthropic2026claudeopus48}, Gemini 3.1 Pro~\citep{deepmind2026gemini31pro}, ChatGPT 5.5~\citep{openai2026gpt55}, and Kimi K2.5~\citep{kimi2026k25}. A candidate is removed as overly easy when all four responses are fully correct under its task-specific scorer.

A retained candidate must have at least one failed screening response. All-model failures still require independent confirmation of sufficient evidence, a stable reference, and an unambiguous output; apparent difficulty alone does not justify acceptance.

\noindent \textbf{Controlled augmentation.}
English and Chinese questions are paired with both document views from Section~\ref{sec:benchmark_contract}. Each variant is checked for semantic equivalence and answerability while preserving the reference, output contract, and scoring criteria. Translated or reformatted variants are not counted as independently authored questions.

\noindent \textbf{Reliability of semantic evaluation.}
We audit GPT-5.4-mini~\citep{openai2026gpt54mini} judgments on 100 responses spanning the four settings, 11 models, and all seven capability groups. A human expert reviews anonymized responses without access to model identities or GPT judgments. With incorrect, partial, and correct ratings, human--GPT agreement reaches 92.0\% and quadratic-weighted Cohen's \(\kappa=0.866\). Including an independent Codex blind review yields ordinal Krippendorff's \(\alpha=0.854\), with a 95\% bootstrap confidence interval of \([0.757,0.924]\). Appendix~\ref{apx:judge_agreement_audit} reports the sampling protocol, score mapping, binary comparison, and review interface.

\section{SciDocIR and SciDocDataset}
\label{sec:ir_data}

\begin{figure}[!tp]
    \centering
    \includegraphics[width=1.0\linewidth]{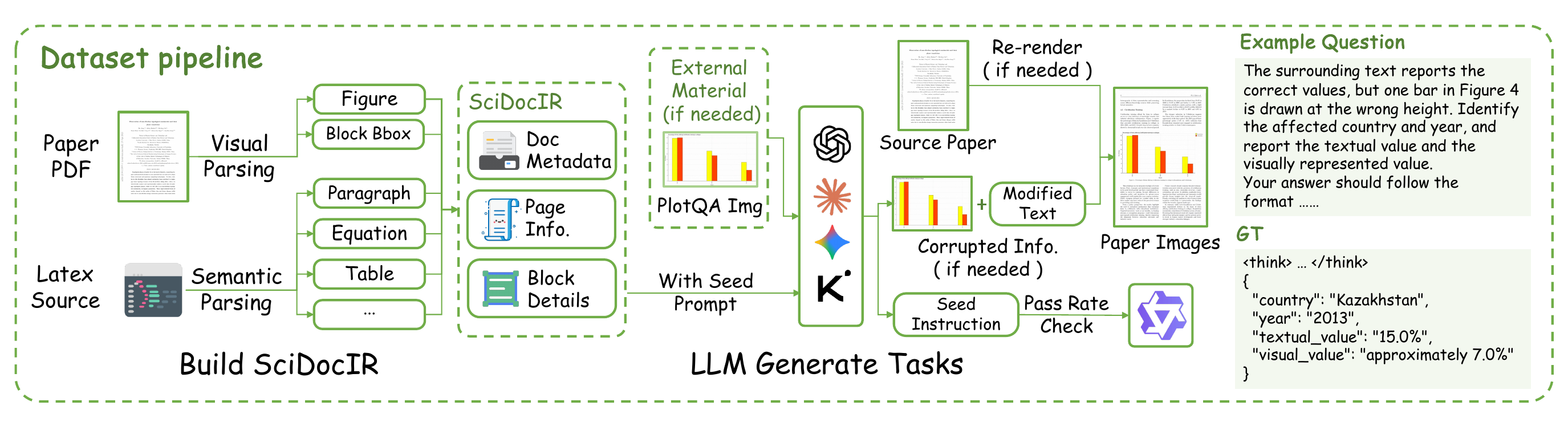}
    \caption{\textbf{\dataset{} construction.} Aligned PDF and LaTeX records form \ir{}. Task-specific generation preserves locked source fields, optionally re-renders controlled evidence into page blocks, and validates instructions and targets against source or construction records.}
    \label{fig:dataset_pipeline}
\end{figure}

Scientific objects can appear as LaTeX environments, rasterized PDF regions, or visually separated blocks. This heterogeneity complicates evidence retrieval, controlled editing, and verification. We normalize papers into \ir{} and construct \dataset{} through verifiable operations over the shared representation (Figure~\ref{fig:dataset_pipeline}).

\subsection{SciDocIR Construction}

PDF pages provide the rendered view: visual parsing identifies text, figures, equations, tables, and captions, recording page indices, bounding boxes, categories, crops, and spatial relations. Available LaTeX sources provide semantic content, section structure, labels, and references. We align these views through text, labels, caption associations, and geometry, preserving both published appearance and source semantics.

Records have three levels: \textbf{document metadata} stores identity, domain, and provenance; \textbf{page information} stores indices, dimensions, images, and layout; \textbf{block details} store content, source code, geometry, captions, reading order, and references. Relations connect continuations, parent--child structures, caption targets, and evidence chains. Appendix~\ref{apx:ir_schema} gives the full schema.

\subsection{Verifiable Training-Task Design}

The 19 benchmark subtasks provide broad diagnostic coverage; the 14 training subtasks are selected for verifiability. They cover layout and reading flow, table logic, chart consistency and readout, cross-document table integration and dataset intersection, notation extraction, symbol disambiguation, and citation roles. Answers are checked through \ir{} fields, controlled perturbations, exact operations, numerical tolerances, parsing, or execution. This separation supports scalable supervision without converting held-out benchmark questions into training examples. Appendix~\ref{apx:training_tasks} lists all construction directions.

\subsection{Seeded Generation and Source Authority}

Each seed instruction specifies the operation, source records, output schema, and rejection conditions. Retrieval selects relevant \ir{} pages and blocks; externally sourced or synthetic tasks provide locked values, labels, or formulas with rendered context. GPT naturalizes questions, proposes bounded candidates, or checks semantics and solvability, but cannot alter deterministic fields or construction records. Appendix~\ref{apx:seed_instructions} specifies model roles and ground-truth authority.

Evidence sources vary by task. Real table integration retrieves compatible tables; simulated integration splits a source table into overlapping, column-permuted views. Chart tasks use PlotQA~\citep{methani2020plotqa} series values and controlled Matplotlib~\citep{hunter2007matplotlib} renderings. Dataset intersection uses a lexicon of approximately 50 scientific resources, while notation tasks use formula and layout templates rendered with XeLaTeX. These sources define exact targets; \ir{} adds realistic document context where needed.
\subsection{Controlled Block-Level Re-rendering}

We insert synthetic figures or tables into their recorded \ir{} bounding boxes, updating the caption or nearby statement when required while preserving unrelated content. This retains the paper's layout and scientific context but gives exact control over task evidence. Unmodified tasks retain original pages. Rendering records store the source, block, geometry, inserted content, textual edits, and expected answer for verification.

\subsection{Validation and Output Records}

Task-specific checks reject missing fields, conflicts with locked records, insufficient visual evidence, and inconsistent or non-executable outputs. Accepted samples contain the instruction, document input, and verifiable answer; source identifiers, locked fields, rendering logs, and rejection outcomes remain in separate audit metadata. This preserves a traceable link between model-assisted wording and answer authority.

\noindent\textbf{Task-aligned training release.}
\dataset{} contains 3,924 SFT instances from 981 seeds under the four language--representation settings, and 10,143 RL instances combining these questions with 6,219 additional verifiable QA instances. It emphasizes evidence localization, structured extraction, numerical verification, and reusable scientific outputs. Appendix~\ref{app:tasks} reports data statistics; Appendix~\ref{apx:posttraining} details the training configurations used in the Qwen3.6 experiments.

\section{Experiments}
\label{sec:exp}

We evaluate 13 proprietary and open multimodal models on all 496 matched \ours{} instances, reporting overall, capability-group, and setting-level scores. We then test whether task-aligned post-training improves scientific-document understanding while preserving general document performance. Appendix~\ref{apx:posttraining} provides experimental configurations and evaluation protocols.

\begin{table}[!tp]
\centering
\caption{\textbf{\ours{} leaderboard.} Overall is computed over all 496 evaluation
instances derived from 124 underlying questions, with failed or unusable responses
assigned a score of zero. Evaluation-setting scores pair English or Chinese questions
with the All Images First (Images) or Markdown Interleaved (Interleaved) document
representation. Capability scores correspond to the seven groups defined in
Section~\ref{sec:benchmark}. Model families are ordered by their best overall score,
and models within each family are sorted by overall score. In each score column the
best, second, and third values are shaded from dark to light; ties receive the same
shading.}
\label{tab:leaderboard}

\scriptsize
\setlength{\tabcolsep}{3pt}
\renewcommand{\arraystretch}{1.16}
\resizebox{\linewidth}{!}{%
\begin{tabular}{@{} l c cccc ccccccc @{}}
\toprule
\multirow{3}{*}{\hd{Model}} & \multirow{3}{*}{\hd{Overall}}
& \multicolumn{2}{c}{English} & \multicolumn{2}{c}{Chinese}
& \multicolumn{7}{c}{Capability scores} \\
\cmidrule(lr){3-4} \cmidrule(lr){5-6} \cmidrule(lr){7-13}
& & \multirow{2}{*}{Images} & Inter- & \multirow{2}{*}{Images} & Inter- & Document & Information & Evidence & Cross- & Reconstr. & Paper-- & Dataset \\
& & & leaved & & leaved & Perception & Extraction & Verification & Document & \& Exec. & Code & Underst. \\
\midrule
\mlogo{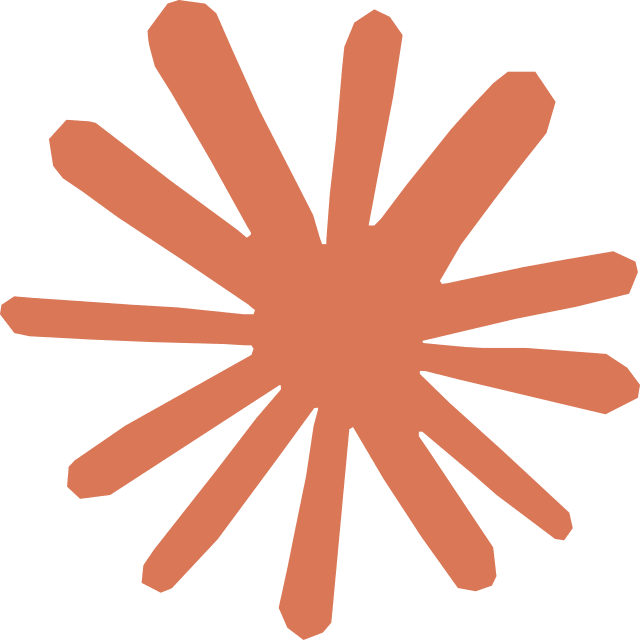}Claude-Opus-5~\citep{anthropic2026claudeopus5} & \rI{62.60} & \rI{63.19} & \rII{61.34} & \rI{65.21} & \rII{60.68} & \rI{61.37} & \rIII{55.52} & \rII{74.47} & \rII{62.81} & \rIII{59.65} & 51.75 & 43.75 \\
\mlogo{claude-color.png}Claude-Opus-4.8~\citep{anthropic2026claudeopus48} & 53.41 & 53.63 & 52.43 & 54.13 & 53.46 & 46.59 & \rI{57.12} & 58.63 & 56.65 & 57.09 & 44.00 & 53.12 \\
\mlogo{claude-color.png}Claude-Sonnet-4.6~\citep{anthropic2026claudesonnet46} & 49.57 & 52.25 & 51.18 & 53.62 & 41.21 & 44.80 & 51.51 & 54.25 & 47.43 & 52.64 & 52.17 & \rIII{59.38} \\
\addlinespace[0.45em]
\mlogo{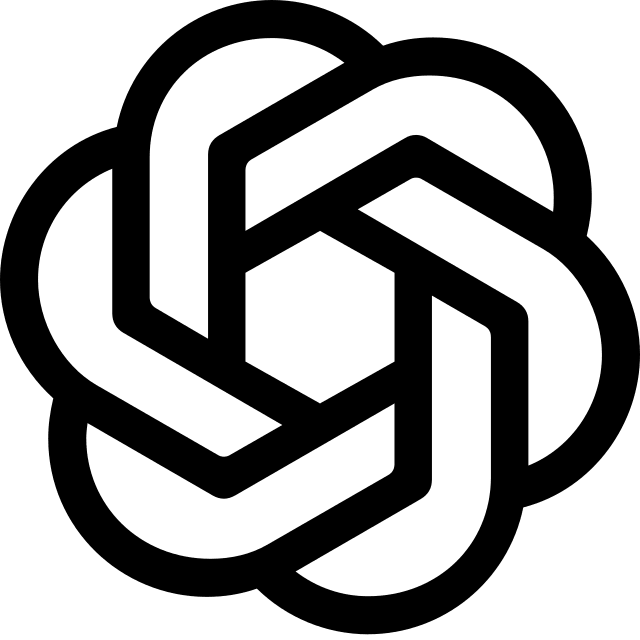}GPT-5.6-Sol~\citep{openai2026gpt56sol} & \rII{61.00} & \rII{61.07} & \rIII{60.32} & \rII{61.32} & \rI{61.31} & \rII{54.08} & \rII{56.78} & \rI{78.16} & 54.62 & \rI{69.70} & 46.69 & \rIII{59.38} \\
\mlogo{openai.png}GPT-5.6-Terra~\citep{openai2026gpt56terra} & 54.19 & 51.88 & 55.69 & 52.83 & 56.37 & 44.01 & 51.74 & 66.28 & 54.01 & \rII{65.75} & \rIII{61.88} & 56.25 \\
\mlogo{openai.png}GPT-5.6-Luna~\citep{openai2026gpt56luna} & 50.58 & 47.55 & 53.99 & 46.65 & 54.13 & 38.68 & 47.52 & 65.15 & 56.07 & 53.86 & 54.25 & \rII{62.50} \\
\addlinespace[0.45em]
\mlogo{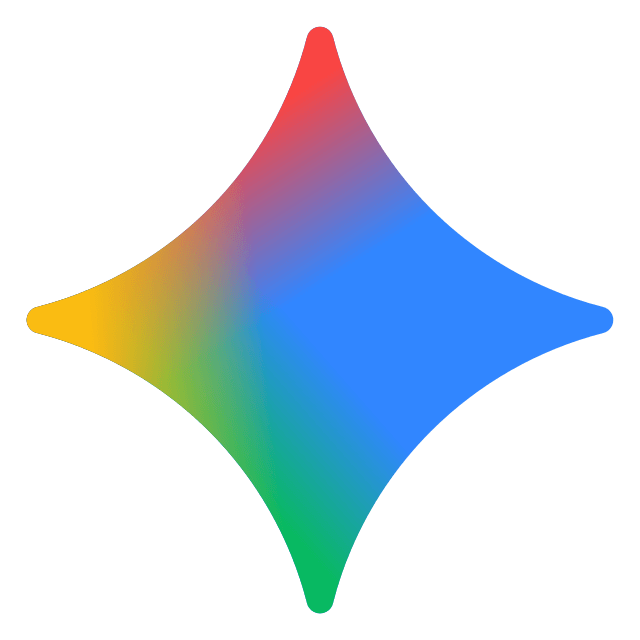}Gemini-3.6-Flash~\citep{deepmind2026gemini36flash} & \rIII{59.87} & 59.19 & \rI{62.98} & \rIII{59.68} & \rIII{57.61} & \rIII{54.00} & 55.19 & 73.28 & \rIII{62.42} & 55.42 & 58.31 & 56.25 \\
\addlinespace[0.45em]
\mlogo{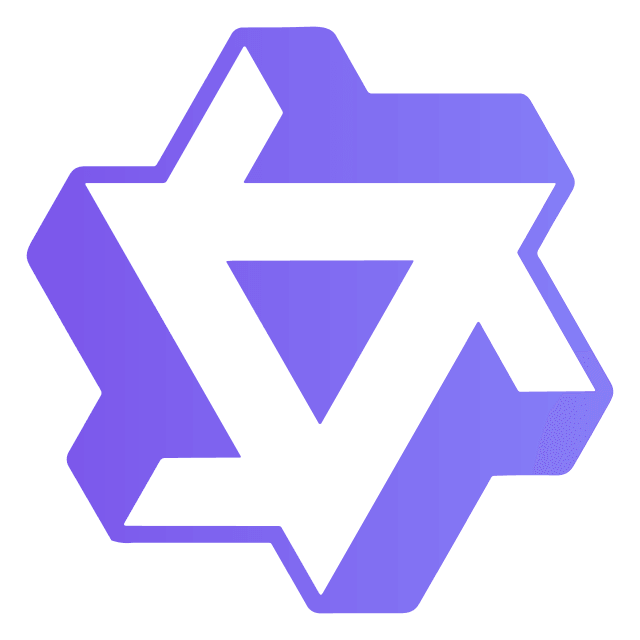}Qwen3.8-Max~\citep{qwen2026qwen38max} & 57.15 & \rIII{59.55} & 55.33 & 58.13 & 55.57 & 48.17 & 50.79 & \rIII{73.84} & \rI{63.02} & 53.60 & 55.75 & \rIII{59.38} \\
\mlogo{qwen-color.png}Qwen3.7-Plus~\citep{qwen2026qwen37plus} & 55.88 & 57.17 & 53.34 & 57.57 & 55.42 & 49.00 & 53.84 & 63.72 & 59.04 & 55.31 & \rII{65.06} & \rI{71.88} \\
\mlogo{qwen-color.png}Qwen3.8-27B~\citep{qwen2026qwen3827b} & 52.89 & 58.65 & 47.25 & 54.21 & 51.46 & 47.43 & 50.50 & 62.87 & 54.80 & 54.45 & 54.06 & 31.25 \\
\addlinespace[0.45em]
\mlogo{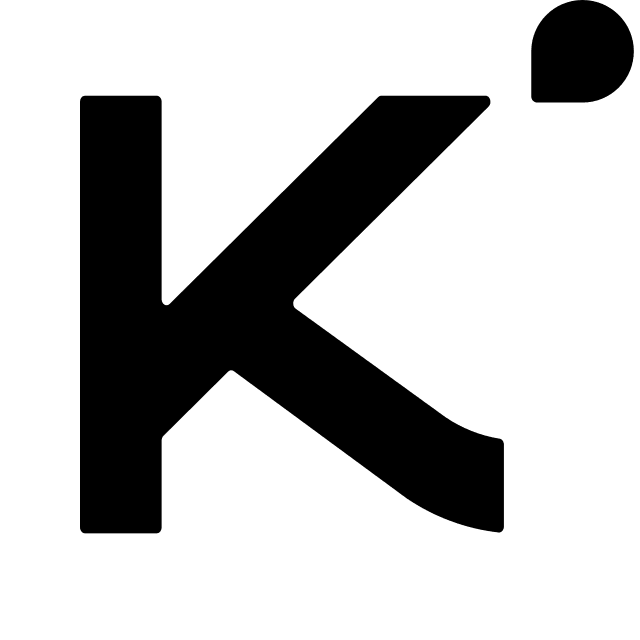}Kimi-K2.5~\citep{kimi2026k25} & 50.38 & 54.74 & 45.42 & 55.04 & 46.33 & 45.69 & 50.62 & 53.05 & 53.61 & 50.41 & \rI{70.25} & 40.62 \\
\addlinespace[0.45em]
\mlogo{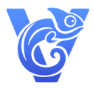}GLM-4.6V~\citep{zai2025glm46v} & 42.44 & 43.30 & 46.70 & 38.21 & 41.56 & 37.15 & 46.58 & 41.63 & 44.57 & 43.90 & 57.27 & \rIII{59.38} \\
\addlinespace[0.45em]
\mlogo{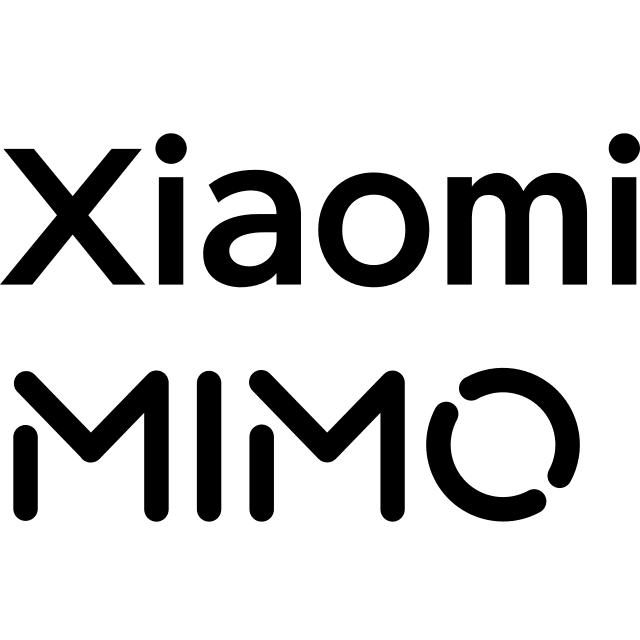}MiMo-V2.5~\citep{xiaomi2026mimov25} & 40.16 & 42.26 & 39.26 & 44.66 & 34.46 & 34.11 & 37.66 & 49.60 & 36.40 & 44.84 & 55.30 & 43.75 \\
\bottomrule
\end{tabular}%
}
\end{table}

\subsection{Overall Leaderboard}

Table~\ref{tab:leaderboard} reports the \ours{} leaderboard. Claude-Opus-5 ranks first with an overall score of 62.60, followed by GPT-5.6-Sol with 61.00 and Gemini-3.6-Flash with 59.87. Qwen3.8-Max and Qwen3.7-Plus obtain 57.15 and 55.88, respectively. No evaluated model reaches 63, and overall scores span 22.44 points across the evaluated systems.

\subsection{Capability-Level Analysis}

\noindent \textbf{Capability leadership is distributed across model families.}
Claude models lead Groups~A and~B, GPT-5.6-Sol leads C and~E, Qwen models lead D and~G, and Kimi-K2.5 leads~F. No model leads more than two groups. Claude-Opus-5's overall lead reflects strength across several groups rather than dominance of the complete taxonomy.

\noindent \textbf{Document perception and scientific extraction form common bottlenecks.}
Averaged equally across the 13 models, Group~A has the lowest score at 46.54, followed by Group~B at 51.18. Group~C has the highest mean at 62.69. The capability ceilings exhibit a related pattern: the best Group~B score is 57.12, while Groups~A and~D peak at 61.37 and 63.02. These results indicate that locating document elements, recovering structured scientific information, and integrating evidence across papers remain difficult even when models perform comparatively well on evidence verification.

\noindent \textbf{High overall performance does not imply a balanced capability profile.}
The gap between each model's strongest and weakest capability groups ranges from 14.58 to 31.62 points. Even Claude-Opus-5 and GPT-5.6-Sol span 30.72 and 31.47 points, with distinct weaknesses in dataset understanding and paper--code alignment, respectively. Overall scores obscure these capability-specific limitations. These contrasting profiles suggest that strong performance on some scientific-document operations does not consistently transfer to others, motivating targeted evaluation and training.

\noindent\textbf{Strong models still lose the scope and completeness of evidence.}
Three Claude-Opus-5 responses illustrate failures beyond answer formatting. Asked for the ordered metrics in a specified section of Depth Anything~3, it includes all four target metrics but prepends two out-of-scope ones, violating the requested list. On a logarithmic plot, it identifies the correct QAOA range but reads the AQC lower endpoint as $5\times10^{-3}$ instead of $8\times10^{-4}$. In a table audit, it catches two planted errors but misses another incorrect mean, returning an extra highlighting complaint instead. These examples point to distinct problems in scope control, scale-aware readout, and exhaustive consistency checking, even when much of the response is locally correct. Appendix~\ref{apx:qualitative_failures} gives the references and predicted outputs.

\subsection{Language and Representation Robustness}

\noindent \textbf{Aggregate language differences conceal model-specific effects.}
Averaged equally over the 13 models and two document representations, English questions score 53.45 and Chinese questions score 52.72, a difference of 0.72 points before rounding. Individual model effects are considerably larger and differ in direction. Qwen3.7-Plus improves by 1.24 points with Chinese questions, whereas GLM-4.6V and Claude-Sonnet-4.6 decrease by 5.12 and 4.30 points, respectively. Language sensitivity is therefore model dependent rather than a uniform translation penalty.

\noindent \textbf{Document representation changes both scores and rankings.}
Across all models and languages, All Images First exceeds Markdown Interleaved by 1.96 points on average. However, nine models favor All Images First and four favor Markdown Interleaved. Kimi-K2.5 and Qwen3.8-27B favor All Images First by 9.02 and 7.07 points, whereas GPT-5.6-Luna favors Markdown Interleaved by 6.96 points. Claude-Opus-5 leads both All Images First settings, Gemini-3.6-Flash leads EN-IL, and GPT-5.6-Sol leads ZH-IL. Neither representation is therefore uniformly preferable across systems.

\noindent \textbf{Language and representation effects can interact.}
For Gemini-3.6-Flash, Markdown Interleaved exceeds All Images First by 3.79 points with English questions, but All Images First exceeds Markdown Interleaved by 2.07 points with Chinese questions. Qwen3.8-27B exhibits a different reversal: Chinese questions reduce its All Images First score by 4.44 points relative to English but improve its Markdown Interleaved score by 4.21 points. These reversals show that the two controlled axes should be analyzed jointly rather than interpreted as independent sources of performance variation.

\noindent \textbf{Cross-setting stability is distinct from peak accuracy.}
The average range across the four evaluation settings is 6.55 points. GPT-5.6-Sol varies by only 1.00 point and Claude-Opus-4.8 by 1.70 points, whereas Claude-Sonnet-4.6, Qwen3.8-27B, and MiMo-V2.5 exhibit ranges above 10 points. Reporting only the best setting would therefore conflate scientific-document capability with compatibility with a particular language and document representation.

\subsection{Post-training with SciDocDataset}
\label{sec:posttraining}

We compare the original Qwen3.6-27B dense model~\citep{qwen2026qwen3627b} with task-aligned SFT and subsequent group relative policy optimization (GRPO)~\citep{shao2024deepseekmath}. These experiments use \dataset{}. Both stages use lightweight LoRA adaptation~\citep{hu2022lora}. Data composition, checkpoint selection, optimization, and reward settings are detailed in Appendix~\ref{apx:posttraining}.

\noindent\textbf{Task-aligned data improves scientific-document performance.}
Table~\ref{tab:posttraining} shows gains of 5.30 points for SFT and 5.71 points for SFT followed by GRPO over the original model, reaching 45.33 and 45.74 versus 40.03. SFT improves all seven capability-group aggregates. Large gains occur in reconstruction and execution (42.28 to 52.48) and paper--code alignment (40.31 to 58.94), where the model must convert document evidence into reusable outputs. Relative to the same baseline, the GRPO-adapted model reaches 61.32 in reconstruction and execution and 46.90 in cross-document understanding, gains of 19.04 and 8.45 points. Both adapted models therefore improve scientific-document performance, with distinct capability profiles.

\begin{table}[!tp]
\centering
\caption{\textbf{Post-training effectiveness and general-document transfer.} Qwen3.6-27B and its adapted variants are evaluated on all 496 \ours{} instances and three general document benchmarks. Input settings follow Table~\ref{tab:leaderboard}. DocVQA and InfoVQA report validation ANLS; ChartQA reports test relaxed accuracy. Scores use final answers only; best scores are bold. Improvement rows give absolute score changes from the original model, computed before rounding; green/red indicates gains/drops.}
\label{tab:posttraining}
\label{tab:general_transfer}
\label{tab:posttraining_capabilities}
\label{tab:posttraining_settings}

\scriptsize
\setlength{\tabcolsep}{1.5pt}
\renewcommand{\arraystretch}{1.16}
\resizebox{\linewidth}{!}{%
\begin{tabular}{@{}l c ccccccc cccc ccc@{}}
\toprule
\multirow{4}{*}{\hd{Model}} & \multicolumn{12}{c}{\ours{}} & \multicolumn{3}{c}{General document understanding} \\
\cmidrule(lr){2-13}\cmidrule(lr){14-16}
 & \multirow{3}{*}{\hd{Overall}} & \multicolumn{7}{c}{Capability scores} & \multicolumn{2}{c}{English} & \multicolumn{2}{c}{Chinese} & \multirow{3}{*}{DocVQA} & \multirow{3}{*}{InfoVQA} & \multirow{3}{*}{ChartQA} \\
\cmidrule(lr){3-9}\cmidrule(lr){10-11}\cmidrule(lr){12-13}
 & & Document & Information & Evidence & Cross- & Reconstr. & Paper-- & Dataset & \multirow{2}{*}{Images} & Inter- & \multirow{2}{*}{Images} & Inter- & & & \\
 & & Perception & Extraction & Verification & Document & \& Exec. & Code & Underst. & & leaved & & leaved & & & \\
\midrule
\mlogo{qwen-color.png}Qwen3.6-27B & 40.03 & 38.14 & 38.55 & 44.71 & 38.46 & 42.28 & 40.31 & 34.38 & 44.61 & 38.61 & 35.38 & 41.53 & 95.81 & 90.67 & 74.60 \\
\nologo{}+ SFT & 45.33 & \textbf{40.11} & \textbf{43.55} & \textbf{51.40} & 43.85 & 52.48 & \textbf{58.94} & \textbf{40.62} & \textbf{46.48} & 42.76 & 45.89 & 46.20 & 95.80 & \textbf{91.02} & 75.40 \\
\rowcolor{scidocorange}
\nologo{}\textit{Improvement} & \gain{+5.30} & \gain{+1.97} & \gain{+5.01} & \gain{+6.69} & \gain{+5.39} & \gain{+10.20} & \gain{+18.62} & \gain{+6.25} & \gain{+1.87} & \gain{+4.15} & \gain{+10.51} & \gain{+4.67} & \loss{-0.01} & \gain{+0.36} & \gain{+0.80} \\
\nologo{}+ SFT + GRPO & \textbf{45.74} & 39.18 & 42.77 & 50.43 & \textbf{46.90} & \textbf{61.32} & 57.10 & \textbf{40.62} & 44.00 & \textbf{44.00} & \textbf{47.38} & \textbf{47.59} & \textbf{95.87} & 90.89 & \textbf{78.00} \\
\rowcolor{scidocorange}
\nologo{}\textit{Improvement} & \gain{+5.71} & \gain{+1.04} & \gain{+4.22} & \gain{+5.73} & \gain{+8.45} & \gain{+19.04} & \gain{+16.79} & \gain{+6.25} & \loss{-0.62} & \gain{+5.39} & \gain{+12.00} & \gain{+6.06} & \gain{+0.05} & \gain{+0.23} & \gain{+3.40} \\
\bottomrule
\end{tabular}%
}
\end{table}

\noindent\textbf{Supervised adaptation transfers across the four input settings.}
SFT improves both languages and both document representations, rather than only the best-performing interface. The gain ranges from 1.87 points on English All Images First to 10.51 points on Chinese All Images First. Averaged over languages, Markdown Interleaved improves from 40.07 to 44.48 after SFT and to 45.80 after GRPO. The matched variants therefore provide useful supervision for expressing the same scientific operation under different input conditions. Full setting-level results are given in Table~\ref{tab:posttraining_settings}.

\noindent\textbf{Producing a final deliverable is part of the task.}
The evaluation permits internal reasoning but scores only the final response against the requested answer contract. Under the same 32,768-token generation budget, the number of instances with a nonempty extracted final answer increases from 323 of 496 for the original model to 461 after SFT and 459 after GRPO. This improvement is relevant to the workflow setting: a model must finish its analysis and return the requested table, structured record, code, or evidence-grounded answer. The training examples pair scientific evidence with explicit output requirements, helping the model translate its analysis into a usable deliverable.

\subsection{Transfer to General Document Benchmarks}
\label{sec:general_transfer}

We evaluate all three models on DocVQA~\citep{mathew2021docvqa}, InfographicVQA~\citep{mathew2022infographicvqa}, and ChartQA~\citep{masry2022chartqa}, without training on their general-document splits. Table~\ref{tab:general_transfer} shows that DocVQA remains near 95.8 across the three models. Relative to the original model's 90.67 on InfoVQA, SFT and SFT followed by GRPO obtain 91.02 and 90.89. With a shared native VLMEvalKit prompt and the recommended Qwen sampling settings, ChartQA scores reach 75.40 and 78.00 versus 74.60 for the original model, gains of 0.80 and 3.40 points.

The ChartQA gains are consistent with the chart readout, numerical verification, and structured extraction tasks in the training collection. Together, the results show that \dataset{} supports scientific-document specialization while preserving general document question answering.

\section{Conclusion}

We introduced \ours{}, a workflow-centered benchmark for scientific document understanding with 124 questions across seven capability groups, 19 subtasks, and four matched evaluation settings. The strongest model scores 62.60, capability leaders differ across model families, and performance depends on document representation. We further introduced \ir{} and \dataset{} to connect these diagnostics with verifiable training supervision. Task-aligned SFT and SFT followed by GRPO improve Qwen3.6-27B over the original model by 5.30 and 5.71 points on \ours{}, while preserving DocVQA and InfoVQA performance and improving ChartQA by 0.80 and 3.40 points. Scientific document understanding is not OCR, long-document QA, or figure reading in isolation: evidence grounding, structured extraction, verification, reconstruction, and provenance tracking must produce outputs that can be checked and reused in scientific work.

\bibliographystyle{plainnat}
\bibliography{references}

\clearpage
\appendix
\section{Limitations}
\label{sec:limit}

Several limitations should be noted. First, the current benchmark contains 124 underlying questions. Although they are expert-authored and difficulty-screened, a larger question pool would strengthen statistical reliability and permit finer-grained capability analysis. The 496 evaluation instances are four matched variants of these questions and should not be interpreted as 496 independent scientific problems. Second, domain coverage emphasizes computer science, mathematics, and selected natural and biomedical sciences; social sciences, humanities, and several engineering fields remain underrepresented. Third, cross-document tasks currently involve two to four papers, while real literature reviews may span dozens of documents. Fourth, synthetic evidence enables precise control and verification but cannot reproduce every visual convention or failure mode found in naturally occurring papers. Finally, \ir{} depends on layout parsing, source alignment, and OCR components. Errors in these upstream stages can propagate into both task generation and training-data quality.

\section{Full SciDocBench Task Taxonomy and Data Statistics}
\label{app:tasks}
Table~\ref{tab:bench_stats} summarizes \textsc{SciDocBench} and \dataset{} .
The source documents are drawn from three channels:
arXiv (primary), academic journals
(including \textit{Nature}, \textit{Journal of the American Chemical Society},
\textit{Chemistry of Materials}, \textit{Organic Letters},
\textit{Biomimetics}, \textit{ACS} family journals, \textit{Global Journal of Environmental Science and Management} and so on),
and web-crawled sources (university repositories and Google Scholar).
 
\begin{table}[!tp]
\centering
\caption{\textbf{Benchmark and training-data statistics.} Training-data totals include validation. Four views of one SFT seed remain in the same split; SFT and RL share the matched-view questions.}
\label{tab:bench_stats}
\footnotesize
\setlength{\tabcolsep}{4pt}
\renewcommand{\arraystretch}{0.98}
\begin{tabularx}{\linewidth}{@{}p{0.53\linewidth}Y@{}}
\toprule
\rowcolor{scidocgreen}
\textbf{Statistic} & \textbf{Value} \\
\midrule
\multicolumn{2}{l}{\textit{Data Source}} \\
\quad Source channels          & arXiv, bioRxiv, OpenReview, PubMed \\
\quad Source papers            & 116 \\
\midrule
\multicolumn{2}{l}{\textit{Benchmark Scale}} \\
\quad Underlying questions     & 124 \\
\quad Evaluation instances     & 496 \\
\quad Scientific domains       & 5 \\
\quad Task categories          & 7 \\
\quad Subtasks                 & 19 \\
\midrule
\multicolumn{2}{l}{\textit{Document \& Question Characteristics}} \\
\quad Avg.\ images per paper   & 14.2 \\
\quad Avg.\ text tokens per instance & 3.9K \\
\midrule
\multicolumn{2}{l}{\textit{Evaluation}} \\
\quad Evaluation methods       & 3 \\
\midrule
\multicolumn{2}{@{}l}{\textit{Training Release}} \\
\quad SFT instances: train / validation / total & 3,844 / 80 / \textbf{3,924} \\
\quad SFT seeds: train / validation / total & 961 / 20 / 981 \\
\quad Views per SFT seed & 2 languages $\times$ 2 document views \\
\quad RL instances: train / validation / total & 10,056 / 87 / \textbf{10,143} \\
\quad RL matched-view / additional QA instances & 3,924 / 6,219 \\
\quad Unique image assets (SFT and RL union) & 55,603 \\
\quad Train--validation partition & Source-paper-disjoint across SFT and RL \\
\bottomrule
\end{tabularx}
\end{table}

\noindent\textbf{Release composition and counting.}
The SFT collection comprises 981 seeds verified against their sources and reference answers. Their four views are also used in RL alongside 6,219 additional verifiable QA instances (6,212 training; seven validation). Connected source papers remain in one split across both stages. Counts precede model-specific length filtering; augmented views and reused questions are not independent tasks.

\begin{figure}[!tp]
    \centering
    \includegraphics[width=0.35\linewidth,trim=0bp 42bp 746bp 42bp,clip]{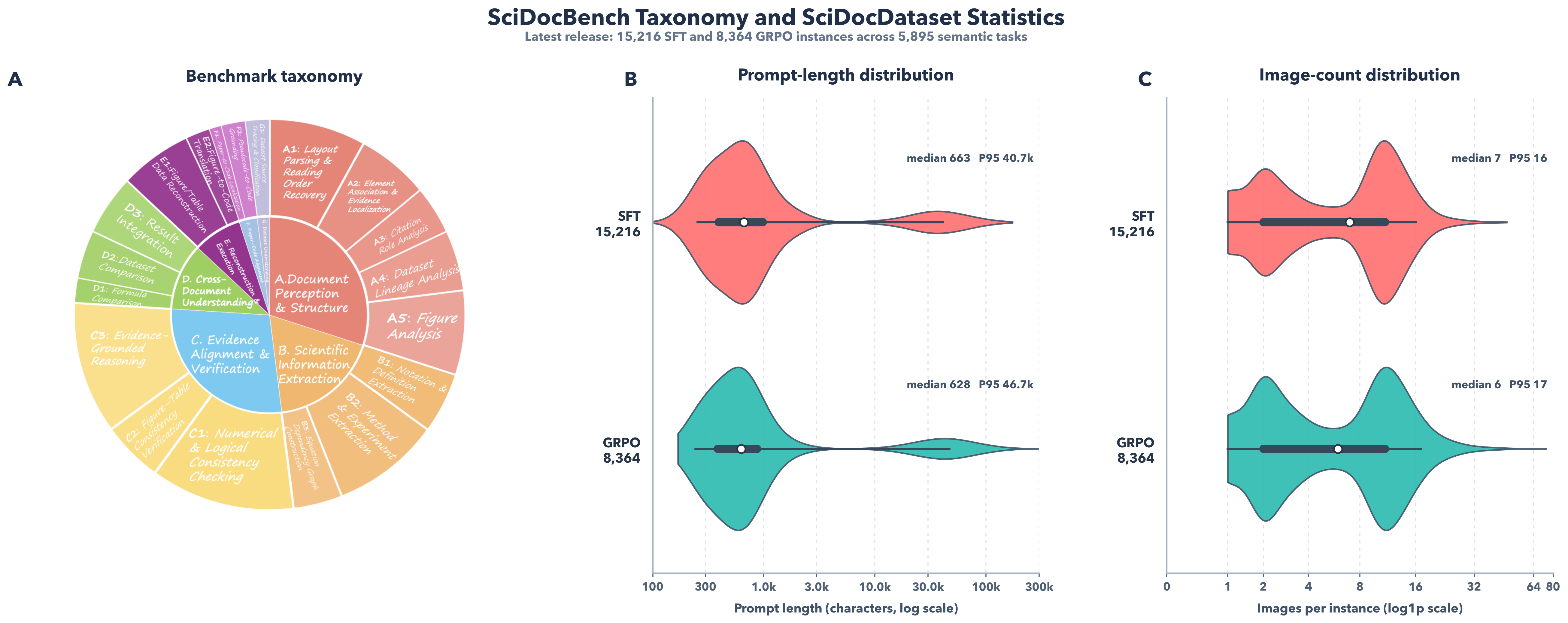}\hfill
    \includegraphics[width=0.64\linewidth]{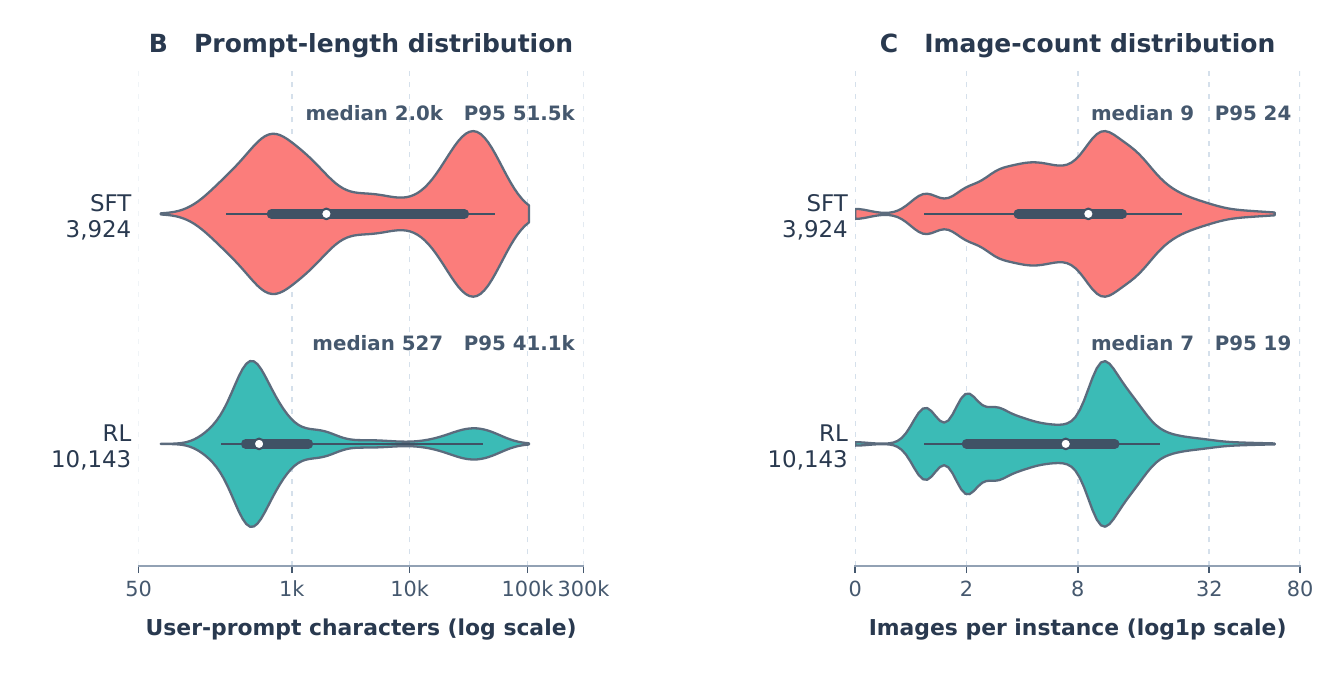}
    \caption{\textbf{\ours{} taxonomy and training-data distributions.} A: benchmark taxonomy. B: user-prompt character counts (log scale). C: image references per instance (log-plus-one scale). Panels B--C cover 3,924 SFT and 10,143 RL instances; dots mark medians and thick bars span the interquartile range.}
    \label{fig:data_overview}
\end{figure}

Figure~\ref{fig:data_overview} shows user-prompt characters (including image placeholders, excluding assistant targets), not tokens, and per-instance image references. Appendix~\ref{apx:posttraining} details the training mixtures and experimental settings.

\begin{table}[!tp]
\centering
\scriptsize
\setlength{\tabcolsep}{3pt}
\renewcommand{\arraystretch}{0.92}
\caption{Full \ours{} task taxonomy. The benchmark contains seven capability groups and nineteen subtasks.}
\label{tab:full_taxonomy}
\begin{tabularx}{\linewidth}{@{}p{0.08\linewidth}p{0.30\linewidth}Y@{}}
\toprule
ID & Task & Evaluation focus \\
\midrule
\multicolumn{3}{@{}l}{\textbf{A. Document Perception \& Structure}} \\
A1 & Layout Parsing \& Reading Order Recovery & Recover local reading flow in two-column PDFs with floating figures, footnotes, and cross-page continuation. \\
A2 & Element Association \& Evidence Localization & Given a claim, locate its supporting figure, table, equation, appendix, page, or block. \\
A3 & Citation Role Analysis & Identify whether cited work acts as background, method basis, data resource, baseline, result support, critique, or extension. \\
A4 & Dataset Lineage Analysis & Recover the source, composition, filtering, derivation, and transformation chain of datasets. \\
A5 & Figure Analysis & Interpret scientific figures, subfigures, legends, axes, annotations, and visual evidence supporting figure-level claims. \\
\midrule
\multicolumn{3}{@{}l}{\textbf{B. Scientific Information Extraction}} \\
B1 & Notation \& Definition Extraction & Build notation indices containing symbols, types, definitions, contexts, and first appearances. \\
B2 & Method \& Experiment Extraction & Extract datasets, metrics, baselines, instruments, configurations, hyperparameters, and experiment stages. \\
B3 & Equation Dependency Graph Construction & Convert mathematical derivations into equation nodes and dependency edges. \\
\midrule
\multicolumn{3}{@{}l}{\textbf{C. Evidence Alignment \& Verification}} \\
C1 & Numerical \& Logical Consistency Checking & Verify whether claims match table values, figure values, rankings, percentages, and derived quantities. \\
C2 & Figure--Table Consistency Verification & Detect whether plotted positions, labels, captions, and tables express the same scientific fact. \\
C3 & Evidence-Grounded Reasoning & Answer questions using document evidence while avoiding unsupported free-form reasoning. \\
\midrule
\multicolumn{3}{@{}l}{\textbf{D. Cross-Document Understanding}} \\
D1 & Cross-Document Formula Comparison & Compare formula variants, assumptions, and generalization paths across papers. \\
D2 & Cross-Document Dataset Comparison & Identify dataset intersections, source overlap, or near-duplicate data usage across papers. \\
D3 & Cross-Document Result Integration & Merge results from multiple papers into a canonical schema with provenance. \\
\midrule
\multicolumn{3}{@{}l}{\textbf{E. Reconstruction \& Execution}} \\
E1 & Figure/Table Data Reconstruction & Recover data points, structured tables, or plotting code from figures and tables. \\
E2 & Figure-to-Code Translation & Convert scientific figures, diagrams, architectures, or flowcharts into runnable or structured code. \\
\midrule
\multicolumn{3}{@{}l}{\textbf{F. Paper--Code Alignment}} \\
F1 & Paper-to-Code Localization & Locate modules, scripts, or functions in a repository corresponding to paper descriptions. \\
F2 & Pseudocode-to-Code Grounding & Complete implementation details by aligning pseudocode, derivations, and source code. \\
\midrule
\multicolumn{3}{@{}l}{\textbf{G. Dataset Understanding}} \\
G1 & Dataset Source Tracing \& Classification & Infer dataset sources, families, and provenance classes from fields, structures, samples, and source patterns. \\
\bottomrule
\end{tabularx}
\end{table}

\subsection{Evaluation Protocol and Per-Question Assignment}
\label{apx:scoring_protocol}

\noindent\textbf{Routing and final-answer scope.}
The released scorer assigns 57 underlying questions to deterministic rules (R), 66 to a semantic LLM judge (J), and one to execution (X), corresponding to 228, 264, and four evaluation instances. Assignment is fixed by question ID and shared by all four views. The audited ID-to-rule mapping takes precedence over the \texttt{eval\_method} metadata field: a JSON output requirement alone does not imply exact matching. Unlisted questions use the semantic judge unless explicitly designated for execution. The source-document tables below give R/J/X for every question; the accompanying \texttt{question\_evaluators.csv} maps each printed ID to its dataset question ID and exact scorer.

Only the final response is evaluated. Delimited internal reasoning is removed; an unfinished reasoning segment, absent final answer, or unusable response receives zero. Explanations requested as part of the final deliverable are still assessed. All evaluator outputs lie in $[0,1]$; no failed instance is dropped from aggregation.

\noindent\textbf{Deterministic rules (R).}
Rules are restricted to questions with closed targets and checkable requirements. Thirty-eight questions compare parsed JSON recursively. Scalars are normalized for benign case, spacing, numeric notation, and location/equation aliases. A dictionary is compared by its specified reference fields, assigning zero to missing fields and normalizing by the larger field count. Lists are compared positionally only when the task requires order; otherwise entries are paired one-to-one by their structured similarity. The list score is normalized by the larger list length, so extra entries are not free. Small unordered lists use maximum-weight matching; lists with more than 12 predicted entries use greedy matching. Parsing failures and incompatible types receive zero. These comparisons concern fixed values or labels, not semantic equivalence of arbitrary prose or LaTeX.

The other 19 rule-scored questions use closed rubrics summarized in Table~\ref{tab:rule_details}. All constants come from the question's scoring contract rather than being inferred from response style. For example, the eight-row indexed-table task awards 0.04 for a valid schema and 0.12 per exact row; the source-tracing task adds 0.33 per correct record and subtracts 0.20 per incorrect record, clipping the result to $[0,1]$. The two keyed-list tasks use schema credits of 0.10 and 0.09 and per-key similarity weights of 0.10 and 0.07, respectively.

\begin{table}[!tp]
\centering
\caption{\textbf{Closed-rule scoring beyond recursive JSON comparison.} These 19 questions and the 38 recursive-JSON questions comprise the 57 R assignments.}
\label{tab:rule_details}
\footnotesize
\setlength{\tabcolsep}{4pt}
\renewcommand{\arraystretch}{1.10}
\begin{tabularx}{\linewidth}{@{}p{0.25\linewidth}cX@{}}
\toprule
\textbf{Rule} & \textbf{Questions} & \textbf{Credit and failure conditions} \\
\midrule
Reading-order ranking & 5 & Fraction of correctly placed labels; incorrect sequence length scores zero. \\
Binary decisions & 6 & Fraction of correct bits in a final bitstring of the required length. \\
Closed scalar & 1 & Normalized label match, with the task's answer-field/label extraction. \\
Grouped choices & 1 & Four groups weighted 0.30, 0.10, 0.30, 0.30; disallowed choices invalidate the affected group. \\
Unordered exact records & 1 & Fraction of reference dictionaries recovered exactly, ignoring key order; one-to-one matching prevents duplicate credit. \\
Indexed exact records & 1 & Exact row count and keys are prerequisites; schema credit plus per-row exact-match credit. \\
Source-tracing records & 1 & Correct-record credit minus incorrect-record penalties; wrong fields invalidate the response. \\
Subfigure and choices & 1 & Wrong subfigure scores zero; otherwise 0.50 plus up to 0.50 for valid choices. \\
Keyed lists & 2 & Valid dictionary-of-lists schema plus weighted structured similarity for each required key. \\
\bottomrule
\end{tabularx}
\end{table}

\noindent\textbf{Semantic judgment (J).}
GPT-5.4-mini receives the question, reference answer, candidate final answer, and task-specific rubric, with temperature zero. It returns a bounded score and a brief assessment. The rubric specifies required content, component weights, prerequisite conditions, and deductions where applicable; the default rubric allocates partial credit by required factual content and penalizes contradictions. Semantically equivalent wording is accepted. JSON key order, surrounding code fences, and harmless whitespace are ignored unless explicitly material to the task. Required fields, evidence relations, units, ordering, and requested explanations remain assessable. Thus open-ended definitions, explanations, and semantically equivalent code or LaTeX are not automatically assigned a byte-exact rule. The judge compares against the vetted reference rather than independently re-reading the full source documents. Responses are cached by judge configuration and grading inputs to reuse identical judgments.

\noindent\textbf{Execution (X).}
The sole execution task, \texttt{constraintfollow\_001}, asks for an image-transformation program. The evaluator extracts Python code and runs it and the reference script on the same input image, with a 30-second timeout per script. Missing output, execution failure, or incorrect dimensions scores zero. Equal output pixels score 1; otherwise mean squared pixel error at most 10 scores 0.5, and larger error scores zero. Pixel comparison uses the reference color mode. Other code-like outputs are not implicitly executed: they retain their listed R or J assignment. Generated programs should be evaluated in an isolated environment.

\subsection{Question Sources and Evaluators}
In the following tables, R denotes deterministic rules, J semantic judgment, and X execution. IDs identify questions, not papers; several questions can share one source. Their dataset question IDs are supplied in the companion evaluation manifest.

\begin{table}[!tp]
\centering
\caption{Source documents and evaluators for Categories A1 to A2.}
\label{tab:A1-A2}

\footnotesize
\setlength{\tabcolsep}{5.0pt}
\renewcommand{\arraystretch}{1.12}
\resizebox{\linewidth}{!}{%
\begin{tabular}{@{}ccccc@{}}
\toprule
Subtask & ID & Subject & Source Document(s) & Eval. \\
\midrule
\multirow{9}{*}{A1}
& 1 & Computer Science & \textit{MM-IFEngine: Towards Multimodal Instruction Following} \cite{ding2025mmifengine} & J \\
\cmidrule(lr){2-5}
& 7 & Nuclear Theory & \makecell[c]{\textit{Revisiting p-$^{11}$B Fusion: Updated Cross-sections,} \\
\textit{Reactivity, and Energy Balance} \cite{wang2026revisiting}} & R \\
\cmidrule(lr){2-5}
& 13 & Computer Science & \makecell[c]{\textit{Depth Anything 3: Recovering the Visual Space} \\
\textit{from Any Views} \cite{lin2026depth}} & R \\
\cmidrule(lr){2-5}
& 51 & Computer Science & \textit{MM-IFEngine: Towards Multimodal Instruction Following} \cite{ding2025mmifengine} & R \\
\cmidrule(lr){2-5}
& 52 & Computer Science & \makecell[c]{\textit{Spatial-SSRL: Enhancing Spatial Understanding} \\
\textit{via Self-Supervised Reinforcement Learning} \cite{liu2025spatial}} & R \\
\cmidrule(lr){2-5}
& 53 & Medicine & \makecell[c]{\textit{QuarkMedSearch: A Long-Horizon Deep Search Agent} \\
\textit{for Exploring Medical Intelligence} \cite{lin2026quarkmedsearch}} & R \\
\cmidrule(lr){2-5}
& 54 & Computer Science & \makecell[c]{\textit{Hyperparameter Transfer Laws for} \\
\textit{Non-Recurrent Multi-Path Neural Networks} \cite{wu2026hyperparameter}} & R \\
\cmidrule(lr){2-5}
& 55 & Computer Science & \makecell[c]{\textit{DocSeeker: Structured Visual Reasoning with Evidence} \\
\textit{Grounding for Long Document Understanding} \cite{yan2026docseeker}} & R \\
\cmidrule(lr){2-5}
& 115 & Computer Science & \makecell[c]{\textit{Depth Anything 3: Recovering the Visual Space} \\
\textit{from Any Views} \cite{lin2026depth}} & J \\
\midrule
\multirow{7}{*}{A2}
& 2 & Computer Science & \textit{LLaVA-CoT: Let Vision Language Models Reason Step-by-Step} \cite{xu2025llava} & J \\
\cmidrule(lr){2-5}
& 8 & Environment & \makecell[c]{\textit{Removal of arsenic (III) and arsenic (V) on chemically} \\
\textit{modified low-cost adsorbent: batch and column operations} \\
\cite{roy2013removal}} & R \\
\cmidrule(lr){2-5}
& 14 & Computer Science & \makecell[c]{\textit{Depth Anything 3: Recovering the Visual Space} \\
\textit{from Any Views} \cite{lin2026depth}} & J \\
\cmidrule(lr){2-5}
& 56 & Computer Science & \makecell[c]{\textit{Booststep: Boosting mathematical capability of large language} \\
\textit{models via improved single-step reasoning} \cite{zhang2025booststep}} & J \\
\cmidrule(lr){2-5}
& 76 & Physics & \makecell[c]{\textit{Cascade of electronic transitions in} \\
\textit{magic-angle twisted bilayer graphene} \cite{wong2020cascade}} & J \\
\cmidrule(lr){2-5}
& 86 & Biology & \makecell[c]{\textit{Inferring the internal structure of groups through the} \\
\textit{integration of statistical learning and causal reasoning} \\
\cite{davis2026inferring}} & R \\
\cmidrule(lr){2-5}
& 106 & Psychology & \makecell[c]{\textit{Inferring the internal structure of groups through the} \\
\textit{integration of statistical learning and causal reasoning} \\
\cite{davis2026inferring}} & R \\
\bottomrule
\end{tabular}}
\end{table}

\begin{table}[!tp]
\centering
\caption{Source documents and evaluators for Categories A3 to A4.}
\label{tab:A3-A4}

\footnotesize
\setlength{\tabcolsep}{5.0pt}
\renewcommand{\arraystretch}{1.12}
\resizebox{\linewidth}{!}{%
\begin{tabular}{@{}ccccc@{}}
\toprule
Subtask & ID & Subject & Source Document(s) & Eval. \\
\midrule
\multirow{9}{*}{A3}
& 3 & Computer Science & \textit{Proximal Policy Optimization Algorithms} \cite{schulman2017proximal} & R \\
\cmidrule(lr){2-5}
& 9 & Biology & \makecell[c]{\textit{ProtFlow: Flow Matching-based Protein Sequence Design} \\
\textit{with Comprehensive Protein Semantic Distribution Learning} \\
\textit{and High-quality Generation} \cite{kong2026protflow}} & R \\
\cmidrule(lr){2-5}
& 15 & Computer Science & \makecell[c]{\textit{Spa3R: Predictive Spatial Field Modeling} \\
\textit{for 3D Visual Reasoning} \cite{jiang2026spa3r}} & R \\
\cmidrule(lr){2-5}
& 77 & Physics & \makecell[c]{\textit{Quantum computing: A taxonomy, systematic review} \\
\textit{and future directions} \cite{gill2022quantum}} & R \\
\cmidrule(lr){2-5}
& 101 & Physics & \makecell[c]{\textit{Quantum computing: A taxonomy, systematic review} \\
\textit{and future directions} \cite{gill2022quantum}} & J \\
\cmidrule(lr){2-5}
& 103 & Computer Science & \makecell[c]{\textit{Quantum computing: A taxonomy, systematic review} \\
\textit{and future directions} \cite{gill2022quantum}} & R \\
\cmidrule(lr){2-5}
& 114 & Biology & \makecell[c]{\textit{ProtFlow: Flow Matching-based Protein Sequence Design} \\
\textit{with Comprehensive Protein Semantic Distribution Learning} \\
\textit{and High-quality Generation} \cite{kong2026protflow}} & J \\
\cmidrule(lr){2-5}
& 119 & Computer Science & \makecell[c]{\textit{Spa3R: Predictive Spatial Field Modeling} \\
\textit{for 3D Visual Reasoning} \cite{jiang2026spa3r}} & J \\
\cmidrule(lr){2-5}
& 121 & Computer Science & \textit{Proximal Policy Optimization Algorithms} \cite{schulman2017proximal} & J \\
\midrule
\multirow{7}{*}{A4}
& 10 & Biology & \makecell[c]{\textit{SaProt: Protein Language Modeling} \\
\textit{with Structure-Aware Vocabulary} \cite{su2023saprot}} & R \\
\cmidrule(lr){2-5}
& 16 & Computer Science & \textit{MM-IFEngine: Towards Multimodal Instruction Following} \cite{ding2025mmifengine} & R \\
\cmidrule(lr){2-5}
& 17 & Computer Science & \makecell[c]{\textit{ChartAssistant: A Universal Chart Multimodal Language} \\
\textit{Model via Chart-to-Table Pre-training and Multitask Instruction Tuning} \\
\cite{meng2024chartassistant}} & R \\
\cmidrule(lr){2-5}
& 57 & Computer Science & \makecell[c]{\textit{ACE-Brain-0: Spatial Intelligence as a Shared} \\
\textit{Scaffold for Universal Embodiments} \cite{gong2026ace}} & R \\
\cmidrule(lr){2-5}
& 87 & Biology & \makecell[c]{\textit{Compressing the collective knowledge of ESM into a} \\
\textit{single protein language model} \cite{dinh2026compressing}, \\
\textit{ProtFlow: Flow Matching-based Protein Sequence Design} \\
\textit{with Comprehensive Protein Semantic Distribution Learning} \\
\textit{and High-quality Generation} \cite{kong2026protflow}, \\
\textit{PTM-Mamba: a PTM-aware protein language model} \\
\textit{with bidirectional gated Mamba blocks} \cite{peng2024ptm}} & R \\
\cmidrule(lr){2-5}
& 107 & Biology & \makecell[c]{\textit{Compressing the collective knowledge of ESM into a} \\
\textit{single protein language model} \cite{dinh2026compressing}, \\
\textit{ProtFlow: Flow Matching-based Protein Sequence Design} \\
\textit{with Comprehensive Protein Semantic Distribution Learning} \\
\textit{and High-quality Generation} \cite{kong2026protflow}} & R \\
\cmidrule(lr){2-5}
& 117 & Computer Science & \makecell[c]{\textit{MM-IFEngine: Towards Multimodal Instruction Following} \\
\cite{ding2025mmifengine}} & J \\
\bottomrule
\end{tabular}}
\end{table}

\begin{table}[!tp]
\centering
\caption{Source documents and evaluators for Categories A5 to B1.}
\label{tab:A5-B1}

\footnotesize  
\setlength{\tabcolsep}{5.2pt}
\renewcommand{\arraystretch}{1.12}
\resizebox{\linewidth}{!}{%
\begin{tabular}{@{}ccccc@{}}
\toprule
Subtask & ID & Subject & Source Document(s) & Eval. \\
\midrule
\multirow{9}{*}{A5}
& 58 & Biology & \makecell[c]{\textit{Estimation of the biological affinities of seven}\\ \textit{species of Sulawesi macaques based on} \\ \textit{multivariate analysis of dermatoglyphic pattern types} \cite{suryobroto1992estimation}} & J \\
\cmidrule(lr){2-5}
& 78 & Chemistry & \makecell[c]{\textit{A Delocalized Mixed-Valence Dinuclear Ytterbium Complex} \\ \textit{That Displays Intervalence Charge Transfer} \cite{obey2024delocalized}} & J \\
\cmidrule(lr){2-5}
& 79 & Computer Science & \makecell[c]{\textit{Biologically Plausible Learning via} \\ \textit{Bidirectional Spike-Based Distillation} \cite{lv2025biologically}} & R \\
\cmidrule(lr){2-5}
& 80 & Physics & \makecell[c]{\textit{Design of a variable stiffness} \\ \textit{quasi-direct drive cable-actuated tensegrity robot} \cite{mi2025design}} & R \\
\cmidrule(lr){2-5}
& 88 & Biology & \makecell[c]{\textit{Deconstruction of a spino-brain--spinal cord} \\ \textit{ circuit that drives chronic pain} \cite{wang2026deconstruction}} & J \\
\cmidrule(lr){2-5}
& 89 & Physics & \makecell[c]{\textit{New limits on the Pauli forbidden transitions in 12C} \\ \textit{nuclei obtained with the complete Borexino dataset} \cite{basilico2026new}} & R \\
\cmidrule(lr){2-5}
& 90 & Physics & \textit{Statics of integrated origami and tensegrity systems}  \cite{ma2023statics} & R \\
\cmidrule(lr){2-5}
& 104 & Computer Science & \makecell[c]{\textit{Biologically Plausible Learning via} \\
\textit{Bidirectional Spike-Based Distillation} \cite{lv2025biologically}} & J \\
\cmidrule(lr){2-5}
& 108 & Physics & \makecell[c]{\textit{New limits on the Pauli forbidden transitions in} \\
\textit{${}^{12}$C nuclei obtained with the complete Borexino dataset} \\
\cite{basilico2026new}} & J \\
\midrule
\multirow{8}{*}{B1}
& 4 & Computer Science & \textit{Group Sequence Policy Optimization} \cite{zheng2025group} & R \\
\cmidrule(lr){2-5}
& 11 & Physics & \makecell[c]{\textit{Coherent spectroscopy with a} \\ \textit{single antiproton spin} \cite{latacz2025coherent}} & J \\
\cmidrule(lr){2-5}
& 12 & Maths & \makecell[c]{\textit{Large time behavior of weak solutions to } \\ \textit{the inhomogeneous incompressible} \\ \textit{Navier-Stokes-Vlasov equations in R3} \cite{su2024large}} & J \\
\cmidrule(lr){2-5}
& 18 & Computer Science & \makecell[c]{ \textit{Hyperparameter Transfer Laws for} \\ \textit{Non-Recurrent Multi-Path Neural Networks} \cite{wu2026hyperparameter}} & R \\
\cmidrule(lr){2-5}
& 81 & Computer Science & \makecell[c]{ \textit{Packed-ensembles for efficient uncertainty estimation} \cite{laurent2022packed}} & J \\
\cmidrule(lr){2-5}
& 102 & Computer Science & \makecell[c]{ \textit{Packed-ensembles for efficient uncertainty estimation} \cite{laurent2022packed}} & J \\
\cmidrule(lr){2-5}
& 116 & Computer Science & \makecell[c]{\textit{Hyperparameter Transfer Laws for Non-Recurrent} \\
\textit{Multi-Path Neural Networks} \cite{wu2026hyperparameter}} & J \\
\cmidrule(lr){2-5}
& 118 & Computer Science & \makecell[c]{\textit{ChartAssistant: A Universal Chart Multimodal Language} \\
\textit{Model via Chart-to-Table Pre-training and Multitask Instruction Tuning} \\
\cite{meng2024chartassistant}} & J \\
\bottomrule
\end{tabular}}
\end{table}

\begin{table}[!tp]
\centering
\caption{Source documents and evaluators for Category B2.}
\label{tab:B2}

\footnotesize
\setlength{\tabcolsep}{5.2pt}
\renewcommand{\arraystretch}{1.12}
\resizebox{\linewidth}{!}{%
\begin{tabular}{@{}ccccc@{}}
\toprule
Subtask & ID & Subject & Source Document(s) & Eval. \\
\midrule
\multirow{10}{*}{B2}
& 5 & Computer Science & \makecell[c]{\textit{Spatial-SSRL: Enhancing Spatial Understanding} \\ \textit{via Self-Supervised Reinforcement Learning} \cite{liu2025spatial}} & R \\
\cmidrule(lr){2-5}
& 19 & Medicine & \makecell[c]{\textit{Glycosylation of anti-dsDNA IgG correlates with} \\ \textit{organ involvement in treatment-naive} \\ \textit{patients with systemic lupus erythematosus} \cite{zhou2025glycosylation}} & R \\
\cmidrule(lr){2-5}
& 20 & Biology & \makecell[c]{\textit{LLM-assisted systematic review } \\ \textit{of large language models in clinical medicine} \cite{chen2026llm}} & J \\
\cmidrule(lr){2-5}
& 21 & Biology & \textit{In vivo site-specific engineering to reprogram T cells} \cite{nyberg2026vivo} & J \\
\cmidrule(lr){2-5}
& 59 & Maths & \makecell[c]{\textit{Quantum linear system solver based on} \\ \textit{time-optimal adiabatic quantum computing} \\ \textit{and quantum approximate optimization algorithm} \cite{an2022quantum}} & J \\
\cmidrule(lr){2-5}
& 69 & Chemistry & \makecell[c]{\textit{Axial chirality-induced rigidification in aminoboranes} \\ \textit{enhances persistent room-temperature phosphorescence} \\ \textit{and circularly polarized luminescence} \cite{eyyathiyil2025axial}} & J \\
\cmidrule(lr){2-5}
& 82 & Chemistry & \makecell[c]{\textit{A Delocalized Mixed-Valence Dinuclear Ytterbium Complex} \\ \textit{That Displays Intervalence Charge Transfer} \cite{obey2024delocalized}} & J \\
\cmidrule(lr){2-5}
& 91 & Physics & \makecell[c]{\textit{A symmetric three degree of freedom tensegrity } \\ \textit{mechanism with dual operation modes for robot actuation} \cite{wang2021symmetric}} & R \\
\cmidrule(lr){2-5}
& 92 & Physics & \makecell[c]{\textit{Impact-resistant, autonomous robots}  \\ \textit{inspired by tensegrity architecture } \cite{johnson2025impact} } & R \\
\cmidrule(lr){2-5}
& 113 & Computer Science & \makecell[c]{\textit{Spatial-SSRL: Enhancing Spatial Understanding via} \\
\textit{Self-Supervised Reinforcement Learning} \cite{liu2025spatial}} & J \\
\bottomrule
\end{tabular}}
\end{table}

\begin{table}[!tp]
\centering
\caption{Source documents and evaluators for Categories B3 to C2.}
\label{tab:B3-C2}

\footnotesize  
\setlength{\tabcolsep}{5.2pt}
\renewcommand{\arraystretch}{1.12}
\resizebox{\linewidth}{!}{%
\begin{tabular}{@{}ccccc@{}}
\toprule
Subtask & ID & Subject & Source Document(s) & Eval. \\
\midrule
\multirow{5}{*}{B3}
& 6 & Computer Science & \textit{Proximal Policy Optimization Algorithms} \cite{schulman2017proximal} & J \\
\cmidrule(lr){2-5}
& 22 & Physics & \makecell[c]{\textit{Observation of non-Abelian topological} \\ \textit{acoustic semimetals and their phase transitions} \cite{jiang2021experimental}} & J \\
\cmidrule(lr){2-5}
& 23 & Biology & \makecell[c]{\textit{Poet: A generative model of } \\ \textit{protein families as sequences-of-sequences} \cite{truong2023poet}} & J \\
\cmidrule(lr){2-5}
& 26 & Maths & \makecell[c]{\textit{Weighted $L^2 $ theory for the Euclidean} \\ \textit{Dirac operator in higher dimensions} \cite{ren2026weighted}} & J \\
\cmidrule(lr){2-5}
& 70 & Maths & \makecell[c]{\textit{Weighted $L^2 $ theory for the Euclidean} \\ \textit{Dirac operator in higher dimensions} \cite{ren2026weighted}} & J \\
\midrule
\multirow{12}{*}{C1}
& 24 & Computer Science & \makecell[c]{\textit{Spatial-SSRL: Enhancing Spatial Understanding} \\ \textit{via Self-Supervised Reinforcement Learning} \cite{liu2025spatial}} & J \\
\cmidrule(lr){2-5}
& 27 & Maths & \makecell[c]{\textit{Learning compositional functions with } \\ \textit{transformers from easy-to-hard data} \cite{wang2025learning}} & J \\
\cmidrule(lr){2-5}
& 28 & Maths & \textit{Dynamic metastability in the self-attention model} \cite{geshkovski2024dynamic} & J \\
\cmidrule(lr){2-5}
& 60 & Environment & \makecell[c]{\textit{Defluoridation of groundwater using aluminum-coated} \\ \textit{bauxite: optimization of synthesis} \\ \textit{process conditions and equilibrium study} \cite{salifu2016defluoridation}} & J \\
\cmidrule(lr){2-5}
& 61 & Computer Science & \textit{Exact recovery in the stochastic block model} \cite{abbe2015exact} & R \\
\cmidrule(lr){2-5}
& 62 & Maths & \textit{Gaussian differential privacy} \cite{dong2022gaussian} & R \\
\cmidrule(lr){2-5}
& 71 & Computer Science & \makecell[c]{\textit{Spa3R: Predictive Spatial Field} \\ \textit{Modeling for 3D Visual Reasoning} \cite{jiang2026spa3r}} & J \\
\cmidrule(lr){2-5}
& 72 & Computer Science & \textit{A unified perspective on the dynamics of deep transformers} \cite{castin2025unified} & R \\
\cmidrule(lr){2-5}
& 73 & Computer Science & \textit{Transformers, parallel computation, and logarithmic depth} \cite{sanford2024transformers} & R \\
\cmidrule(lr){2-5}
& 74 & Maths & \textit{Benign overfitting in linear regression} \cite{bartlett2020benign} & R \\
\cmidrule(lr){2-5}
& 83 & Physics & \makecell[c]{\textit{Magic-angle graphene superlattices:} \\ \textit{a new platform for unconventional superconductivity} \cite{cao2018magic}} & J \\
\cmidrule(lr){2-5}
& 84 & Computer Science & \makecell[c]{\textit{Provable failure of language models in learning majority} \\ \textit{ boolean logic via gradient descent} \cite{chen2025provable}} & R \\
\midrule
\multirow{5}{*}{C2}
& 25 & Computer Science & \makecell[c]{\textit{Spatial-SSRL: Enhancing Spatial Understanding} \\ \textit{via Self-Supervised Reinforcement Learning} \cite{liu2025spatial}} & J \\
\cmidrule(lr){2-5}
& 29 & Physics & \makecell[c]{\textit{Gaia data release 3-summary} \\ \textit{of the content and survey properties} \cite{vallenari2023gaia}} & J \\
\cmidrule(lr){2-5}
& 63 & Computer Science & \makecell[c]{\textit{Scalecap: Inference-time scalable image captioning} \\ \textit{via dual-modality debiasing} \cite{xing2025scalecap}} & J \\
\cmidrule(lr){2-5}
& 64 & Biology & \makecell[c]{\textit{Neural sequences underlying directed } \\ \textit{turning in Caenorhabditis elegans} \cite{kramer2026neural}} & J \\
\cmidrule(lr){2-5}
& 93 & Computer Science & \textit{Improved baselines with visual instruction tuning} \cite{liu2024improved} & R \\
\bottomrule
\end{tabular}}
\end{table}

\begin{table}[!tp]
\centering
\caption{Source documents and evaluators for Categories C3 to D1.}
\label{tab:C3-D1}

\footnotesize  
\setlength{\tabcolsep}{5.2pt}
\renewcommand{\arraystretch}{1.12}
\resizebox{\linewidth}{!}{%
\begin{tabular}{@{}ccccc@{}}
\toprule
Subtask & ID & Subject & Source Document(s) & Eval. \\
\midrule
\multirow{10}{*}{C3}
& 30 & Medicine & \makecell[c]{\textit{Glycosylation of anti-dsDNA IgG correlates with} \\ \textit{organ involvement in treatment-naive} \\ \textit{patients with systemic lupus erythematosus} \cite{zhou2025glycosylation}} & R \\
\cmidrule(lr){2-5}
& 31 & Environment & \makecell[c]{\textit{Simultaneous removal of heavy metals from aqueous } \\ \textit{solutions by pineapple crown and } \\ \textit{avocado peel hydrogel composites} \cite{wardani2026simultaneous}} & J \\
\cmidrule(lr){2-5}
& 32 & Maths & \makecell[c]{\textit{The Euler Stratification for } \textit{$\mathbb{P}^1 \times \mathbb{P}^1 \times \mathbb{P}^n$} \cite{hocsten2026euler}} & J \\
\cmidrule(lr){2-5}
& 33 & Maths & \textit{Computation and sampling for Schubert specializations} \cite{anderson2026computation} & R \\
\cmidrule(lr){2-5}
& 34 & Maths & \makecell[c]{\textit{Learning compositional functions with } \\ \textit{transformers from easy-to-hard data} \cite{wang2025learning}} & J \\
\cmidrule(lr){2-5}
& 66 & Biology & \makecell[c]{\textit{Neural sequences underlying directed turning} \\ \textit{in Caenorhabditis elegans} \cite{kramer2026neural}} & R \\
\cmidrule(lr){2-5}
& 75 & Biology & \makecell[c]{\textit{Deconstruction of a spino-brain--spinal cord} \\ \textit{ circuit that drives chronic pain} \cite{wang2026deconstruction}} & J \\
\cmidrule(lr){2-5}
& 85 & Chemistry & \makecell[c]{\textit{Giant coercivity and high magnetic blocking temperatures for N2$^{3-}$} \\ \textit{ radical-bridged dilanthanide complexes upon ligand dissociation} \cite{demir2017giant}} & J \\
\cmidrule(lr){2-5}
& 94 & Physics & \makecell[c]{\textit{A tensegrity-inspired bidirectional quasi-zero } \\ \textit{stiffness metamaterial for buffering and energy absorption} \cite{shen2025tensegrity}} & J \\
\cmidrule(lr){2-5}
& 105 & Chemistry & \makecell[c]{\textit{Giant coercivity and high magnetic blocking temperatures} \\
\textit{for N$_2^{3-}$ radical-bridged dilanthanide complexes} \\
\textit{upon ligand dissociation} \cite{demir2017giant}} & J \\
\midrule
\multirow{3}{*}{D1}
& 35 & Computer Science & \makecell[c]{\textit{Proximal Policy Optimization Algorithms} \cite{schulman2017proximal},\\
\textit{Deepseekmath: Pushing the limits of} \\ \textit {mathematical reasoning in open language models} \cite{shao2024deepseekmath}, \\
\textit{Group Sequence Policy Optimization} \cite{zheng2025group},\\
\textit{Minimax-m1: Scaling test-time} \\ \textit{compute efficiently with lightning attention} \cite{chen2025minimax}} & R \\
\cmidrule(lr){2-5}
& 65 & Physics & \makecell[c]{\textit{Approximating quantum many-body wave}\\ \textit{functions using artificial neural networks} \cite{cai2018approximating},\\
\textit{Quantum entanglement in neural network states} \cite{deng2017quantum},\\
\textit{Solving the quantum many-body problem}\\ \textit{with artificial neural networks} \cite{carleo2017solving}} & J \\
\cmidrule(lr){2-5}
& 67 & Physics & \makecell[c]{\textit{Approximating quantum many-body wave}\\ \textit{functions using artificial neural networks} \cite{cai2018approximating},\\
\textit{Quantum entanglement in neural network states} \cite{deng2017quantum},\\
\textit{Solving the quantum many-body problem}\\ \textit{with artificial neural networks} \cite{carleo2017solving}} & J \\
\bottomrule
\end{tabular}}
\end{table}

\begin{table}[!tp]
\centering
\caption{Source documents and evaluators for Category D2.}
\label{tab:D2}

\footnotesize  
\setlength{\tabcolsep}{5.2pt}
\renewcommand{\arraystretch}{1.12}
\resizebox{\linewidth}{!}{%
\begin{tabular}{@{}ccccc@{}}
\toprule
Subtask & ID & Subject & Source Document(s) & Eval. \\
\midrule
\multirow{6}{*}{D2}
& 36 & Computer Science & \makecell[c]{\textit{Eagle 2.5: Boosting long-context post-training } \\ \textit {for frontier vision-language models} \cite{chen2025eagle}, \\
\textit{Mammoth-vl: Eliciting multimodal reasoning with }\\
\textit{instruction tuning at scale} \cite{guo2025mammoth}} & J \\
\cmidrule(lr){2-5}
& 37 & Computer Science & \makecell[c]{\textit{Cambrian-1: A fully open, vision-centric} \\ \textit {exploration of multimodal llms} \cite{tong2024cambrian}, \\
\textit{Llava-onevision: Easy visual task transfer} \cite{li2024llava},\\
\textit{Mammoth-vl: Eliciting multimodal reasoning with }\\
\textit{instruction tuning at scale} \cite{guo2025mammoth}} & R \\
\cmidrule(lr){2-5}
& 95 & Chemistry & \makecell[c]{\textit {Strategies for pre-training graph neural networks} \cite{hu2019strategies}, \\
\textit{Geometry-enhanced molecular representation} \\ \textit{learning for property prediction} \cite{fang2022geometry},\\
\textit{Molecular contrastive learning of representations}\\
\textit{ via graph neural networks} \cite{wang2022molecular}} & R \\
\cmidrule(lr){2-5}
& 96 & Biology & \makecell[c]{\textit{ProtFlow: Flow Matching-based Protein Sequence} \\ \textit{Design with Comprehensive Protein Semantic} \\ \textit{Distribution Learning and High-quality Generation} \cite{kong2026protflow},\\
\textit{PTM-Mamba: a PTM-aware protein language model}\\ \textit{with bidirectional gated Mamba blocks} \cite{peng2024ptm},\\\textit{Compressing the collective knowledge of ESM} \\ \textit{ into a single protein language model} \cite{dinh2026compressing}} & R \\
\cmidrule(lr){2-5}
& 109 & Chemistry & \makecell[c]{\textit{Molecular Contrastive Learning of Representations via} \\
\textit{Graph Neural Networks} \cite{wang2022molecular}, \\
\textit{ChemRL-GEM: Geometry Enhanced Molecular Representation} \\
\textit{Learning for Property Prediction} \cite{fang2022geometry}, \\
\textit{Strategies for Pre-Training Graph Neural Networks} \\
\cite{hu2019strategies}} & J \\
\cmidrule(lr){2-5}
& 120 & Computer Science & \makecell[c]{\textit{Cambrian-1: A Fully Open, Vision-Centric Exploration} \\
\textit{of Multimodal LLMs} \cite{tong2024cambrian}, \\
\textit{LLaVA-OneVision: Easy Visual Task Transfer} \cite{li2024llava}, \\
\textit{MAmmoTH-VL: Eliciting Multimodal Reasoning with} \\
\textit{Instruction Tuning at Scale} \cite{guo2025mammoth}} & R \\
\bottomrule
\end{tabular}}
\end{table}

\begin{table}[!tp]
\centering
\caption{Source documents and evaluators for Category D3.}
\label{tab:D3}

\footnotesize
\setlength{\tabcolsep}{5.2pt}
\renewcommand{\arraystretch}{1.12}
\resizebox{\linewidth}{!}{%
\begin{tabular}{@{}ccccc@{}}
\toprule
Subtask & ID & Subject & Source Document(s) & Eval. \\
\midrule
\multirow{8}{*}{D3}
& 38 & Computer Science & \makecell[c]{\textit{SpatialDreamer: Incentivizing Spatial Reasoning} \\ \textit { via Active Mental Imagery} \cite{cao2025spatialdreamer}, \\
\textit{Spatialladder: Progressive training for}\\
\textit{spatial reasoning in vision-language models} \cite{li2025spatialladder}} & J \\
\cmidrule(lr){2-5}
& 39 & Computer Science & \makecell[c]{\textit{Ts-llava: Constructing visual tokens through thumbnail-and-} \\ \textit {sampling for training-free video large language models} \cite{qu2024ts}, \\
\textit{Pllava: Parameter-free llava extension from images }\\
\textit{to videos for video dense captioning} \cite{xu2024pllava}} & J \\
\cmidrule(lr){2-5}
& 47 & Medicine & \makecell[c]{\textit{Glycosylation of anti-dsDNA IgG correlates with} \\ \textit{organ involvement in treatment-naive} \\ \textit{patients with systemic lupus erythematosus} \cite{zhou2025glycosylation},\\
\textit{IgG glycans in health and disease: Prediction,}\\ \textit{intervention, prognosis, and therapy} \cite{shkunnikova2023igg}} & R \\
\cmidrule(lr){2-5}
& 97 & Computer Science & \makecell[c]{\textit{Caprl: Stimulating dense image caption} \\ \textit{ capabilities via reinforcement learning} \cite{xing2025caprl},\\
\textit{Scalecap: Inference-time scalable image captioning} \\ \textit{via dual-modality debiasing} \cite{xing2025scalecap}} & R \\
\cmidrule(lr){2-5}
& 98 & Chemistry & \makecell[c]{\textit{High voltage cycling stability of LiF-coated NMC811 electrode} \cite{llanos2024high}, \\ \textit{Long-term cyclability of NCM-811 at high voltages in}\\
\textit{lithium-ion batteries: an in-depth diagnostic study} \cite{li2020long}, \\ \textit{Enhanced Cycling Stability of NCM811 Cathodes at High}\\ \textit{C-Rates and Voltages via LiMTFSI-Based Polymer Coating} \cite{kim2025enhanced}} & R \\
\cmidrule(lr){2-5}
& 110 & Computer Science & \makecell[c]{\textit{ScaleCap: Inference-Time Scalable Image Captioning via} \\
\textit{Dual-Modality Debiasing} \cite{xing2025scalecap}, \\
\textit{CapRL: Stimulating Dense Image Caption Capabilities via} \\
\textit{Reinforcement Learning} \cite{xing2025caprl}} & R \\
\cmidrule(lr){2-5}
& 111 & Chemistry & \makecell[c]{\textit{Long-Term Cyclability of NCM-811 at High Voltages in} \\
\textit{Lithium-Ion Batteries: an In-Depth Diagnostic Study} \\
\cite{li2020long}, \\
\textit{Enhanced Cycling Stability of NCM811 Cathodes at High} \\
\textit{C-Rates and Voltages via LiMTFSI-Based Polymer Coating} \\
\cite{kim2025enhanced}, \\
\textit{High Voltage Cycling Stability of LiF-Coated NMC811 Electrode} \\
\cite{llanos2024high}} & J \\
\cmidrule(lr){2-5}
& 123 & Medicine & \makecell[c]{\textit{IgG glycans in health and disease: Prediction, intervention,} \\
\textit{prognosis, and therapy} \cite{shkunnikova2023igg}, \\
\textit{Fucosylation of anti-dsDNA IgG1 correlates with disease} \\
\textit{activity of treatment-na\"ive systemic lupus erythematosus patients} \\
\cite{han2022fucosylation}} & J \\
\bottomrule
\end{tabular}}
\end{table}

\begin{table}[!tp]
\centering
\caption{Source documents and evaluators for Categories E, F, and G.}
\label{tab:EFG}

\footnotesize  
\setlength{\tabcolsep}{5.2pt}
\renewcommand{\arraystretch}{1.12}
\resizebox{\linewidth}{!}{%
\begin{tabular}{@{}ccccc@{}}
\toprule
Subtask & ID & Subject & Source Document(s) & Eval. \\
\midrule
\multirow{8}{*}{E1}
& 40 & Computer Science & \textit{Group Sequence Policy Optimization} \cite{zheng2025group} & R \\
\cmidrule(lr){2-5}
& 41 & Computer Science & \textit{LLaVA-CoT: Let Vision Language Models Reason Step-by-Step} \cite{xu2025llava} & J \\
\cmidrule(lr){2-5}
& 48 & Physics & \makecell[c]{\textit{Stairway Codes: Floquetifying Bivariate}\\ \textit{Bicycle Codes and Beyond} \cite{jacoby2026stairway}} & J \\
\cmidrule(lr){2-5}
& 49 & Physics & \makecell[c]{\textit{Imaginary-time evolution of interacting spin}\\ \textit{systems in the truncated Wigner approximation} \cite{schlegel2026imaginary}} & R \\
\cmidrule(lr){2-5}
& 68 & Biology & \makecell[c]{\textit{Neural sequences underlying directed } \\ \textit{turning in Caenorhabditis elegans} \cite{kramer2026neural}} & J \\
\cmidrule(lr){2-5}
& 99 & Biology & \makecell[c]{\textit{Causal modelling of gene effects}\\ \textit{from regulators to programs to traits} \cite{ota2026causal}} & R \\
\cmidrule(lr){2-5}
& 112 & Biology & \makecell[c]{\textit{Causal modelling of gene effects from regulators} \\
\textit{to programs to traits} \cite{ota2026causal}} & J \\
\cmidrule(lr){2-5}
& 124 & Physics & \makecell[c]{\textit{Imaginary-time evolution of interacting spin systems} \\
\textit{in the truncated Wigner approximation} \\
\cite{schlegel2026imaginary}} & J \\
\midrule
\multirow{2}{*}{E2}
& 42 & Computer Science & \textit{MM-IFEngine: Towards Multimodal Instruction Following} \cite{ding2025mmifengine} & X \\
\cmidrule(lr){2-5}
& 50 & Physics & \makecell[c]{\textit{QFlowNet: Fast, Diverse, and Efficient Unitary} \\ \textit{Synthesis with Generative Flow Networks} \cite{koo2026qflownet}} & J \\
\midrule
\multirow{2}{*}{F1}
& 43 & Computer Science & \textit{MM-IFEngine: Towards Multimodal Instruction Following} \cite{ding2025mmifengine} & R \\
\cmidrule(lr){2-5}
& 122 & Computer Science & \makecell[c]{\textit{TwiFF (Think With Future Frames): A Large-Scale Dataset} \\
\textit{for Dynamic Visual Reasoning} \cite{liu2026twiff}, \\
\textit{AgentVista: Evaluating Multimodal Agents in Ultra-Challenging} \\
\textit{Realistic Visual Scenarios} \cite{su2026agentvista}, \\
\textit{MM-IFEngine: Towards Multimodal Instruction Following} \\
\cite{ding2025mmifengine}, \\
\textit{Spatial-SSRL: Enhancing Spatial Understanding via} \\
\textit{Self-Supervised Reinforcement Learning} \cite{liu2025spatial}} & R \\
\midrule
\multirow{2}{*}{F2}
& 44 & Biology & \textit{On the genetic basis of tail-loss evolution in humans and apes} \cite{xia2024genetic} & J \\
\cmidrule(lr){2-5}
& 100 & Biology & \textit{On the genetic basis of tail-loss evolution in humans and apes} \cite{xia2024genetic} & R \\
\midrule
\multirow{2}{*}{G1}
& 45 & Computer Science & \makecell[c]{\textit{MM-IFEngine: Towards Multimodal Instruction Following} \cite{ding2025mmifengine},\\
\textit{Agentvista: Evaluating multimodal agents in}\\ \textit{ultra-challenging realistic visual scenarios} \cite{su2026agentvista},\\
\textit{TwiFF (Think With Future Frames): A Large-Scale}\\ \textit{Dataset for Dynamic Visual Reasoning} \cite{liu2026twiff}, \\
\textit{Spatial-SSRL: Enhancing Spatial Understanding} \\ \textit{via Self-Supervised Reinforcement Learning} \cite{liu2025spatial}} & R \\
\cmidrule(lr){2-5}
& 46 & Biology & \makecell[c]{\textit{Understanding protein function with a multimodal}\\ \textit{retrieval-augmented foundation model} \cite{truong2025understanding},\\
\textit{Progen2: exploring the boundaries of protein language models} \cite{nijkamp2023progen2}} & J \\
\bottomrule
\end{tabular}}
\end{table}

\FloatBarrier
\section{Benchmark Annotation and Verification Protocol}
\label{apx:annotation_verification}

\begin{figure*}[!tp]
    \centering
    \includegraphics[width=0.98\textwidth]{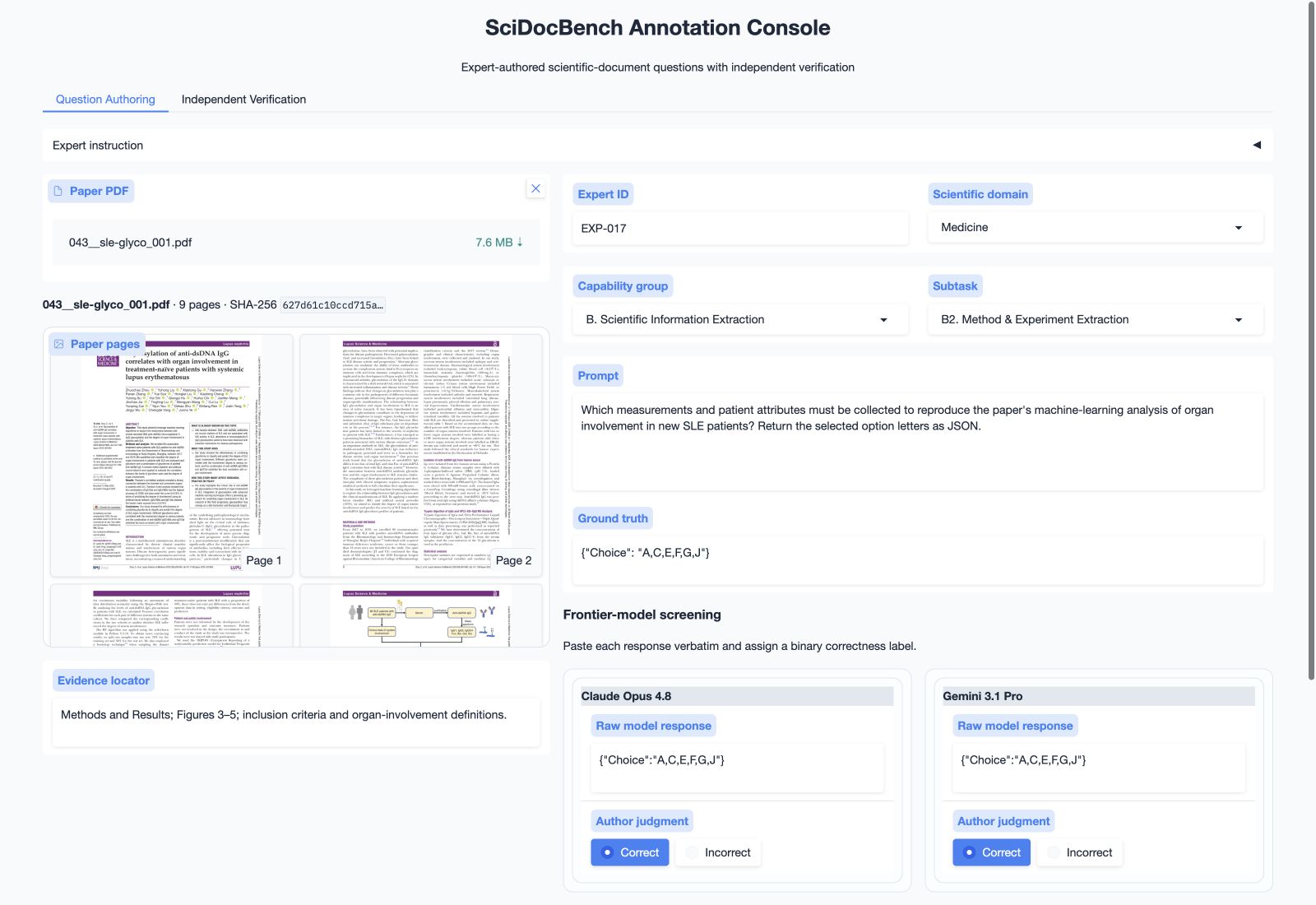}
    \caption{Expert question-authoring interface. The interface presents the source PDF and rendered pages alongside the task metadata, question, ground truth, evidence locator, and verbatim screening-model responses with author-assigned correctness labels. The populated fields constitute a representative interface record used to illustrate the annotation procedure and are not additional leaderboard results.}
    \label{fig:annotation_authoring_ui}
\end{figure*}

Benchmark construction uses two separate interfaces and two annotator roles. The authoring interface records the source PDF, expert identifier, scientific domain, capability group, subtask, finalized question, ground-truth answer, and evidence locator. The evidence locator identifies the page, section, figure, table, equation, appendix, or other document object from which the answer can be verified. Authors then run the finalized item once on each of the four designated screening models, paste the responses without editing, and assign a binary correctness label to every response. An item can be submitted only when all required fields and four screening records are present and at least one screened model is incorrect. If the question, ground truth, or evidence locator changes, the four responses are collected again. Figure~\ref{fig:annotation_authoring_ui} shows the authoring interface.

\begin{center}
\fcolorbox{black!20}{blue!3}{%
\begin{minipage}{0.94\linewidth}
\textbf{Instructions for Question Authors.}
Read the supplied paper and create one scientifically grounded question that is answerable from the paper but challenging for current multimodal models.
\begin{enumerate}[leftmargin=1.4em,itemsep=1pt,topsep=3pt,parsep=0pt]
    \item Upload the source PDF; provide the finalized prompt and ground-truth answer; and identify the page, section, figure, table, equation, appendix, or other scientific object that supports the answer.
    \item Write a clear and self-contained prompt with an unambiguous output format and a stable answer that can be verified directly from the supplied document. Do not rely on unstated external knowledge.
    \item Run the finalized question once on each designated screening model. Paste every response verbatim and label it \textbf{Correct} or \textbf{Incorrect} according to the ground truth. Do not edit, merge, or selectively rerun individual responses.
    \item Submit the item only when every required field and all four screening records are complete and at least one valid screening response is incorrect. If the prompt, ground truth, or evidence locator changes, collect all four responses again.
\end{enumerate}
\end{minipage}%
}
\end{center}

\begin{figure*}[!tp]
    \centering
    \includegraphics[width=0.98\textwidth]{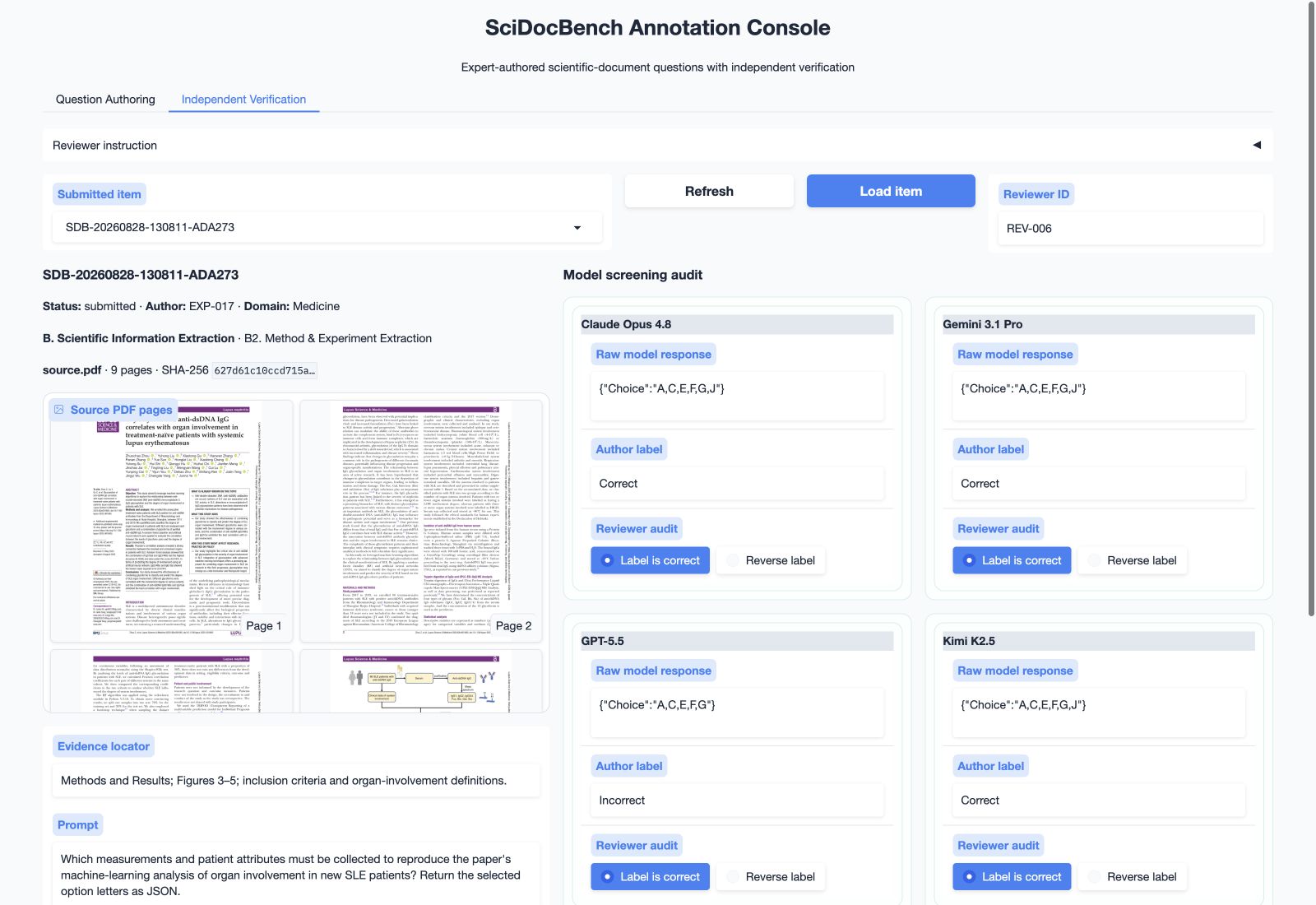}
    \caption{Independent-verification interface. Reviewers inspect the same PDF, task metadata, evidence locator, question, ground truth, and four screening records; audit each correctness label; complete the acceptance checklist; and record an Accept, Revise, or Reject decision.}
    \label{fig:annotation_verification_ui}
\end{figure*}

Every submitted item is subsequently inspected by a reviewer who did not author the question. The reviewer loads the immutable submission and verifies four conditions: the question is answerable from the supplied PDF, the wording and requested output format are unambiguous, the ground truth is correct, and at least one screening model is incorrect. The reviewer also audits every response label and can reverse an author label when it does not agree with the ground truth. An item is accepted only when all four checks pass. Items with correctable defects are returned for revision, while items with unstable evidence or irreparable ambiguity are rejected. Figure~\ref{fig:annotation_verification_ui} shows the independent-verification interface.

The annotation system stores each submission as an atomic record containing a schema version, item identifier, author and reviewer identifiers, timestamps, task metadata, the question and ground truth, the evidence locator, the source-PDF filename and SHA-256 digest, the four verbatim responses, the author labels, reviewer label audits, and the final decision. The PDF digest prevents accidental substitution of the source document after screening. Screening responses are used only to remove items already solved by all designated models; they are not used to construct the ground truth. This separation preserves expert evidence as the basis of correctness while retaining an auditable record of question difficulty and reviewer agreement.

\subsection{LLM-as-a-Judge Reliability Audit}
\label{apx:judge_agreement_audit}

We assess the reliability of semantic scoring on 100 response-level instances drawn from accepted LLM-as-a-judge evaluation runs. The audit set contains 25 instances from each of the four matched evaluation settings, covers 11 evaluated models, and includes questions from all seven capability groups. Sampling is approximately balanced across the judge's initial incorrect, partial, and correct score bands. Deterministic rule-based and execution-based evaluations are excluded because their correctness is established by programmatic scorers rather than semantic judgment.

The human expert was shown only the question, reference answer, and anonymized model response. Model identity, the GPT-5.4-mini score, and the judge rationale were hidden until all 100 ratings had been submitted. The expert assigned one of three scores: 0 for an incorrect response, 0.5 for a partially correct response, and 1 for a fully correct response. For the ordinal comparison, continuous GPT-5.4-mini and Codex scores are mapped using the same strict endpoint rule: 0 is incorrect, 1 is correct, and every value strictly between 0 and 1 is partially correct. We report quadratic-weighted Cohen's $\kappa$~\citep{cohen1968weighted} alongside exact agreement. The audit therefore measures consistency in applying the reference-based scoring rubric; it does not constitute an independent re-annotation of each source document.

\begin{table}[!tp]
\centering
\scriptsize
\caption{Pairwise agreement on the 100-instance semantic-judge audit. Binary results threshold the original scores at 0.5. QWK denotes quadratic-weighted Cohen's $\kappa$, and MAE denotes mean absolute score difference.}
\label{tab:judge_agreement_audit}
\setlength{\tabcolsep}{5.0pt}
\renewcommand{\arraystretch}{1.08}
\begin{tabular}{@{}lccccc@{}}
\toprule
Rater pair & Three-level exact & QWK & Binary exact & Binary $\kappa$ & MAE \\
\midrule
Human expert vs. GPT-5.4-mini & 92.0\% & 0.866 & 70.0\% & 0.355 & 0.169 \\
Codex vs. GPT-5.4-mini & 89.0\% & 0.818 & 80.0\% & 0.593 & 0.128 \\
Human expert vs. Codex & 93.0\% & 0.872 & 74.0\% & 0.405 & 0.145 \\
\bottomrule
\end{tabular}
\end{table}

The average pairwise three-level exact agreement is 91.3\%. To summarize all three rating sources jointly, we compute ordinal Krippendorff's $\alpha$~\citep{hayes2007reliability} over the expert, GPT-5.4-mini, and Codex labels. The resulting $\alpha$ is 0.854, with a 95\% confidence interval of $[0.757, 0.924]$ obtained from 10,000 bootstrap resamples of the audit instances. The stronger ordinal agreement than binary agreement reflects the intended use of partial credit: thresholding at 0.5 converts small differences within the intermediate score range into categorical disagreements.

\begin{figure}[!tp]
    \centering
    \includegraphics[width=0.96\linewidth,height=0.76\textheight,keepaspectratio]{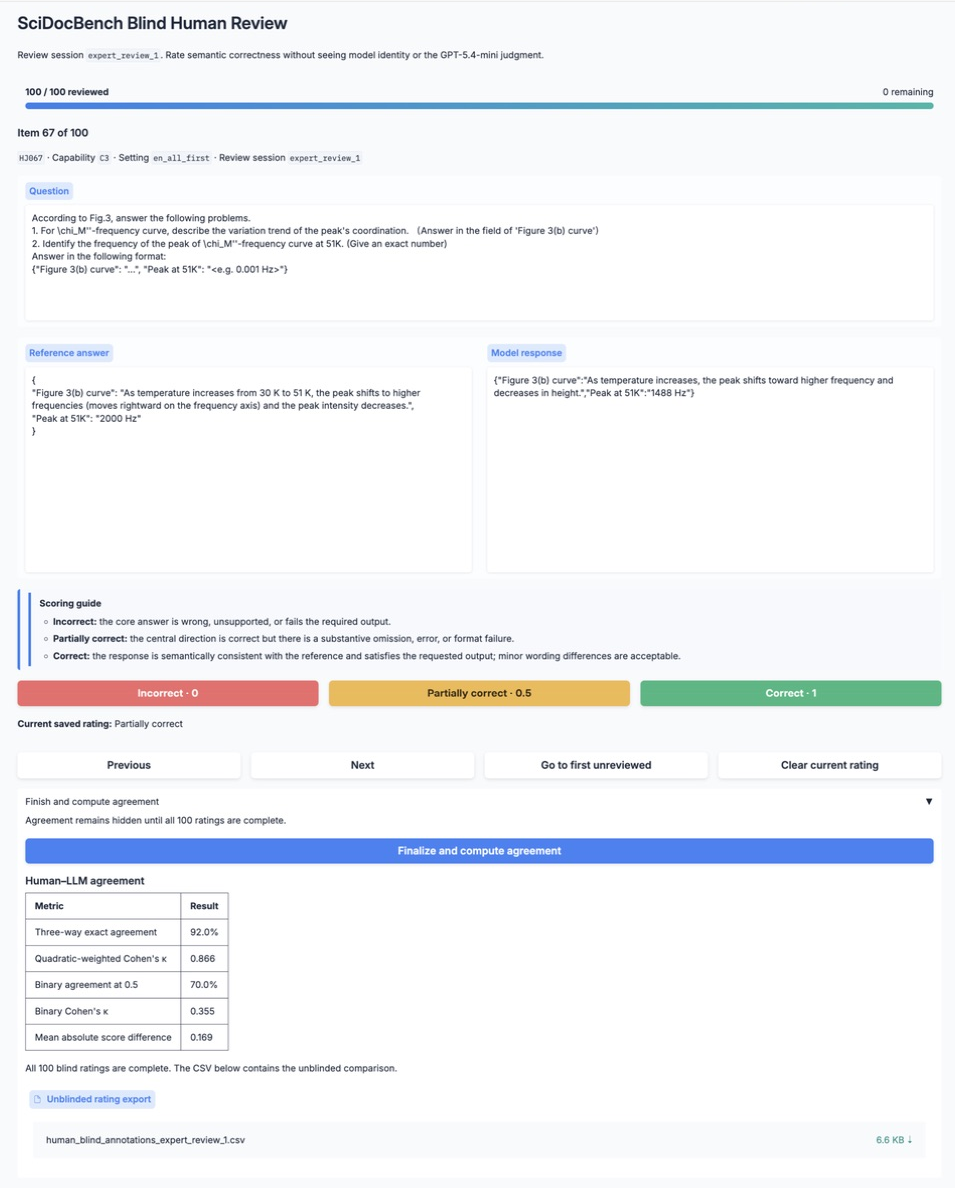}
    \caption{Complete blind expert-review interface for the English audit item HJ067. The interface presents the item metadata, question, reference answer, anonymized model response, three-level rubric, rating controls, navigation controls, and the agreement summary revealed after all 100 ratings were completed. Model identity and the GPT-5.4-mini judgment remain hidden during rating. Browser chrome and unrelated tabs are omitted from the capture.}
    \label{fig:judge_blind_review_ui}
\end{figure}

\FloatBarrier
\section{SciDocIR Schema}
\label{apx:ir_schema}

Figure~\ref{fig:ir_schema} presents the simplified schema of \ir{}. The representation stores document-level metadata, page-level metadata, block-level records, relations, and task-specific metadata. Optional fields are populated only when they are relevant to task generation or reward verification.

\begin{figure}[!tp]
\centering
\begin{minipage}{0.96\linewidth}
\begin{lstlisting}[style=scidocir]
Record SciDocIR
    document  : {doc_id, source_dir, num_pages, provenance}
    categories: List<{id, name, supercategory}>
    pages     : List<Page>
    blocks    : List<Block>
    relations : List<Relation>
    stats     : {num_blocks, num_relations, category_counts}
End Record

Record Page
    page_index : Integer
    width, height : Float
    image_path : String
End Record

Record Block
    block_id, page_index : Integer
    bbox       : {x1, y1, x2, y2, width, height}
    category   : {id, name, supercategory}
    content    : {source_code, format, plain_text, latex}
    structure  : {previous_block, parent_block, next_block}
    references : {labels, outgoing_ref_labels, incoming_ref_from}
    metadata   : Metadata
End Record

Record Metadata
    title_level       : Integer?
    caption_text      : String?
    caption_target    : {type, block_ids}?
    figure_info       : {crop_path, visual_ocr, chart_type, subfigures}?
    table_structure   : TableStructure?
    equation_latex    : String?
    nearby_block_ids  : List<BlockId>
    task_fields       : Dict<String, Any>
End Record

Record Relation
    type : "adj" | "sub" | "peer" | "identical" | "continuation"
    from : BlockId
    to   : BlockId
End Record
\end{lstlisting}
\end{minipage}
\caption{A simplified schema of \ir{}. Optional fields are filled only when useful for generation or verification.}
\label{fig:ir_schema}
\end{figure}

\clearpage
\section{Training-Data Directions and Verifiable Subtasks}
\label{apx:training_tasks}

Table~\ref{tab:training_tasks} lists the eight construction directions and fourteen verifiable subtasks underlying \dataset{}. These describe task construction rather than dataset size; instance and split counts are given in Table~\ref{tab:bench_stats}.

\begin{table}[!tp]
\centering
\small
\caption{Training-data construction directions and their instantiated subtasks.}
\label{tab:training_tasks}
\begin{tabularx}{\linewidth}{@{}p{0.30\linewidth}Y@{}}
\toprule
Direction & Training subtasks \\
\midrule
Layout and reading-flow recovery & Block role classification; parent--child linking; reading-order prediction; next-hop reading-target prediction. \\
Table logic consistency & Table logic consistency check. \\
Cross-document table integration & Real cross-document table merge; single-document simulated table merge. \\
Chart readout and recovery & Single-point chart readout; multi-point or series readout. \\
Chart visual consistency & Chart visual consistency. \\
Cross-document dataset comparison & Cross-document dataset intersection. \\
Notation understanding & Notation extraction; symbol ambiguity disambiguation. \\
Citation-role analysis & A3-lite citation-role classification. \\
\bottomrule
\end{tabularx}
\end{table}

\subsection{Construction Sources and Model Roles}

The seed instruction is not an assertion that every sample is generated end to end by an LLM. It is a task-specific contract governing the part assigned to GPT-5.4~\citep{openai2026gpt54}. Deterministic labels, values, alignments, and perturbation records are locked before model invocation whenever possible. GPT-5.4 is used for linguistic naturalization, bounded candidate generation, semantic classification, or solvability checks. Tables~\ref{tab:seed_sources_a} and~\ref{tab:seed_sources_b} identify the records supplied to each template and the authority used for ground truth.

\begin{table}[!tp]
\centering
\scriptsize
\setlength{\tabcolsep}{2.4pt}
\renewcommand{\arraystretch}{1.05}
\caption{Construction sources and GPT-5.4 roles for subtasks 1--7. ``None'' denotes no external dataset beyond the \ir{} paper collection.}
\label{tab:seed_sources_a}
\begin{tabularx}{\linewidth}{@{}p{0.19\linewidth}p{0.27\linewidth}p{0.18\linewidth}Y@{}}
\toprule
Subtask & \ir{} or document records & External or synthetic source & GPT-5.4 role and GT authority \\
\midrule
Block role classification & Page image; target page index, bounding box, type, and layout annotation & None & Light question rewriting. The layout label remains the locked GT. \\
Parent--child linking & Block text, page and bounding box; heading level; caption pairs; footnote and continuation relations & None & Question rewriting and optional plausibility check. The verified relation remains GT. \\
Reading-order prediction & Order and reading annotations; block pages and boxes; section and discourse roles & None & Question rewriting only. GT follows the stored order and layout rules. \\
Next-hop reading target & Current and successor blocks; caption pairs; figure/table references; cross-page and appendix links & None & Question rewriting and optional next-hop plausibility check. The stored successor remains GT. \\
Table logic consistency & Table LaTeX, normalized JSON, caption, headers, cells, citing context, page, and box & None & Generates a natural corrupted statement, repair, and distractors, then self-checks them. Rules lock the perturbed field and answer. \\
Real cross-document table merge & Table schemas, rows, columns, values, captions, and nearby text from different papers & None & Polishes context and question. Rules align compatible schemas and compute the merged GT. \\
Simulated table merge & One real table and its context, split into two partially overlapping tables with permuted columns & Programmatic table splitting and a simulated second paper page & Adjusts table LaTeX, local prose, and question without changing locked cells. Rules compute the complete merged GT. \\
\bottomrule
\end{tabularx}
\end{table}

\begin{table}[!tp]
\centering
\scriptsize
\setlength{\tabcolsep}{2.4pt}
\renewcommand{\arraystretch}{1.05}
\caption{Construction sources and GPT-5.4 roles for subtasks 8--14.}
\label{tab:seed_sources_b}
\begin{tabularx}{\linewidth}{@{}p{0.19\linewidth}p{0.27\linewidth}p{0.18\linewidth}Y@{}}
\toprule
Subtask & \ir{} or document records & External or synthetic source & GPT-5.4 role and GT authority \\
\midrule
Single-point chart readout & Figure and caption blocks; page, box, and nearby text used for page replacement & PlotQA chart, axes, series labels, and source values & Naturalizes the question and performs repeated visual solvability checks. PlotQA values remain GT. \\
Multi-point or series readout & Same figure placement and context records as the single-point task & Complete PlotQA series data & Rewrites the question and checks the returned list. The source series remains GT. \\
Chart visual consistency & Figure position, caption, and page context used to reinsert a controlled chart & PlotQA values; Matplotlib radar, line, scatter, and bar renderers & Rewrites the question and checks solvability. Rules move one plotted value and retain the original annotation; the corruption log defines GT. \\
Dataset intersection & Dataset-like entities mined from text, tables, and captions when available; simulated two-paper text & Internal lexicon of approximately 50 CS, AI, mathematics, and natural-science resources; alias and composition pools & Checks name plausibility and polishes page prose. Exact set intersection defines GT. \\
Notation extraction & Final batch does not require real \ir{} records & Mathematics, machine-learning, and physics formula templates; XeLaTeX page rendering & May polish surrounding prose while formulas and definitions stay locked. The saved symbol table defines GT. \\
Symbol ambiguity disambiguation & Final batch does not require real \ir{} records & Formula and context templates; one-column, two-column, and spanning layouts & May naturalize local contexts without changing definitions. The rule log for each symbol occurrence defines GT. \\
A3-lite citation role & Block LaTeX and text; real citation keys; section, page, block identifier, and local context & No external bibliography database & Rules propose a role from lexical cues and section context; GPT-5.4 classifies the observed use. Ambiguous or disagreeing cases are rejected. \\
\bottomrule
\end{tabularx}
\end{table}

\subsection{Seed Instruction Templates}
\label{apx:seed_instructions}

The templates below show the shared generation contract and the task-specific seed appended to it. Angle-bracketed expressions are populated by the retrieval, synthesis, or rendering program. A locked field cannot be altered by the model. The model-facing training instance contains the generated question and its associated document input; construction metadata and source identifiers are retained only for generation, verification, and auditing.

\begin{lstlisting}[style=scidocseed]
You construct one scientific-document training instance from supplied records.

Inputs may include:
- IR_RECORDS: selected SciDocIR pages, blocks, relations, and metadata.
- DOCUMENT_VIEW: the page image, crop, or rendered document shown to the learner.
- STRUCTURED_SOURCE: source table, chart data, formula template, or entity lists.
- CONSTRUCTION_RECORD: locked labels, values, mappings, perturbations, and tolerances.

Use only the supplied material. Do not invent a scientific fact, citation, identifier, value, unit, relation, or source. Preserve every field marked LOCKED. Write one self-contained question that is answerable from DOCUMENT_VIEW and that states the required answer format. Do not reveal the answer or hidden construction metadata in the question.

Return JSON only:
{
  "question": "<user-facing instruction>",
  "answer": <task-specific schema>,
  "source_ids": ["<records used for audit>"],
  "quality_check": {"answerable": true, "single_interpretation": true}
}

If a locked gold answer is supplied, copy it exactly. If the task asks for a bounded semantic label, select only from the supplied ontology. Return {"reject": "<reason>"} when the evidence is incomplete, ambiguous, illegible, or inconsistent with the construction record.
\end{lstlisting}

\newcommand{\seedtaskheading}[1]{%
  \Needspace{5\baselineskip}%
  \par\noindent\textbf{#1}\par\nobreak\vspace{0.2em}%
}

\Needspace{5\baselineskip}
\subsubsection{Layout and Reading-Flow Recovery}

\seedtaskheading{1. Block role classification.}
\begin{lstlisting}[style=scidocseed]
Inputs: PAGE_IMAGE, TARGET_BLOCK {page_index, bbox, type}, LAYOUT_ANNOTATION, ROLE_LABELS.

Mark TARGET_BLOCK visibly in the page view. Rewrite the base instruction as a concise classification question asking for exactly one label from ROLE_LABELS. Use nearby layout only as context and do not mention the stored type. Copy LAYOUT_ANNOTATION[TARGET_BLOCK] as the answer. Reject a crop containing several inseparable semantic blocks or a label absent from ROLE_LABELS.

Answer schema: "<one label from ROLE_LABELS>"
\end{lstlisting}

\seedtaskheading{2. Parent--child linking.}
\begin{lstlisting}[style=scidocseed]
Inputs: BLOCK_A, BLOCK_B, local page context, heading levels, caption pairs, footnote links, and continuation relations; LOCKED_RELATION.

Present A and B with visible labels and enough context to determine their structural relation. Ask whether B is the content targeted by A as a caption, a continuation of A, or unrelated. Use exactly caption_target, continuation, and none. Naturalize the wording without changing block contents. Copy LOCKED_RELATION as the answer and reject pairs that require unseen context or admit more than one relation.

Answer schema: "caption_target" | "continuation" | "none"
\end{lstlisting}

\seedtaskheading{3. Reading-order prediction.}
\begin{lstlisting}[style=scidocseed]
Inputs: PAGE_IMAGE, CANDIDATE_BLOCKS with page_index and bbox, ORDER_ANNOTATION, READING_ANNOTATION, and discourse roles; LOCKED_ORDER.

Assign visible labels A, B, C, ... in the supplied shuffled order. Ask for the order in which the selected blocks should be read. Ensure that the view retains columns, headings, captions, and any cross-page cue needed to solve the task. Copy LOCKED_ORDER, expressed as a label sequence, as the answer. Reject cyclic, disconnected, or visually ambiguous selections.

Answer schema: "<sequence such as A>B>D>C>"
\end{lstlisting}

\seedtaskheading{4. Next-hop reading-target prediction.}
\begin{lstlisting}[style=scidocseed]
Inputs: CURRENT_BLOCK, CANDIDATE_BLOCKS, verified successor, caption pairs, textual figure/table references, cross-page relations, and appendix links; LOCKED_NEXT_LABEL.

Label two to six candidates A through F. Ask which block should be read immediately after CURRENT_BLOCK in order to continue the document or follow the explicit reference. Keep plausible distractors from the same local context. Copy LOCKED_NEXT_LABEL as the answer. Reject the sample if two candidates are reasonable next hops or the required link is not visible.

Answer schema: "<one candidate letter>"
\end{lstlisting}

\Needspace{5\baselineskip}
\subsubsection{Table Logic and Integration}

\seedtaskheading{5. Table logic consistency.}
\begin{lstlisting}[style=scidocseed]
Inputs: TABLE_LATEX, TABLE_JSON, caption, headers, cells, nearby citing text, and PERTURBATION_RECORD with one locked change.

Using the locked perturbation, write one natural but false statement about a value, entity, rank, comparison direction, or metric direction. Also write its minimal correction and three plausible statements that are exactly supported by the table. Shuffle the four statements, ask which one is inconsistent, and report the corresponding letter. Independently check every option against TABLE_JSON. Do not create another inconsistency.

Answer schema: {"choice": "A|B|C|D", "repair": "<corrected statement>"}
\end{lstlisting}

\seedtaskheading{6. Real cross-document table merge.}
\begin{lstlisting}[style=scidocseed]
Inputs: TABLE_A and TABLE_B from different papers, captions, nearby text, SCHEMA_MAP, and LOCKED_MERGE.

Write a concise cross-paper question asking the learner to merge the compatible rows and columns under the canonical schema in SCHEMA_MAP. Preserve method names, dataset splits, metric direction, units, and paper provenance exactly. Do not request a comparison that the mapping does not support. Copy LOCKED_MERGE as the answer; rules, not the language model, determine all aligned keys and values.

Answer schema: {"columns": ["..."], "rows": [{"row_key": "...", "values": {...}, "sources": {...}}]}
\end{lstlisting}

\seedtaskheading{7. Single-document simulated table merge.}
\begin{lstlisting}[style=scidocseed]
Inputs: one real SOURCE_TABLE, SPLIT_RECORD, two simulated paper contexts, and LOCKED_MERGE.

The construction program has split SOURCE_TABLE into TABLE_A and TABLE_B with different column orders and partially overlapping rows. Improve the two LaTeX tables and their surrounding prose so they read as independent paper excerpts, but preserve every locked header and cell. Ask for a complete merge with duplicate rows reconciled by SPLIT_RECORD. Copy LOCKED_MERGE as the answer and reject any rewrite that changes a value, unit, or row identity.

Answer schema: {"columns": ["..."], "rows": [{"row_key": "...", "values": {...}, "sources": {...}}]}
\end{lstlisting}

\Needspace{5\baselineskip}
\subsubsection{Chart Readout and Visual Consistency}

\seedtaskheading{8. Single-point chart readout.}
\begin{lstlisting}[style=scidocseed]
Inputs: a PlotQA chart and source data, TARGET_POINT {series, x, value, unit}, target IR figure bbox, caption, and nearby paper text.

After the chart is placed in the paper page, write a natural question identifying one unambiguous point by series and x-axis condition. Ask for its value and unit and state the allowed precision when needed. Copy TARGET_POINT.value and unit as the answer. Perform repeated visual checks that the point, axes, and legend are legible; reject overlapping, clipped, or uncertain points.

Answer schema: {"series": "<name>", "x": "<condition>", "value": <number>, "unit": "<unit>"}
\end{lstlisting}

\seedtaskheading{9. Multi-point or series readout.}
\begin{lstlisting}[style=scidocseed]
Inputs: a PlotQA chart, FULL_SERIES_DATA, target IR figure bbox, caption, and nearby paper text; LOCKED_TARGETS.

Write a question asking for either several specified points or one complete short series in displayed x-axis order. State the unit and numeric precision. Preserve the locked series and x labels exactly. Copy the requested values from FULL_SERIES_DATA and visually verify the complete list after rendering. Reject stacked, occluded, discontinuous, or dual-axis cases without an explicit unambiguous mapping. Prefer single-point conversion if the full series is not reliably readable.

Answer schema: {"series": "<name>", "x": ["..."], "values": [<numbers>], "unit": "<unit>"}
\end{lstlisting}

\seedtaskheading{10. Chart visual consistency.}
\begin{lstlisting}[style=scidocseed]
Inputs: PlotQA source values, Matplotlib chart rendering, IR figure bbox, caption and page context, and CORRUPTION_RECORD for exactly one moved plotted value.

The rendered chart contains one visual error while its label, caption, or surrounding text preserves the correct value. Write a question asking for the affected series or entity, x-condition, correct value, and visually represented value. Do not reveal which mark was moved. Copy all fields from CORRUPTION_RECORD. Check that every other mark is consistent and that both values are visually recoverable within TOLERANCE.

Answer schema: {"entity": "<series or entity>", "condition": "<x-label>", "correct_value": <number>, "visual_value": <number>, "unit": "<unit>"}
\end{lstlisting}

\Needspace{5\baselineskip}
\subsubsection{Cross-Document Dataset Intersection}

\seedtaskheading{11. Cross-document dataset intersection.}
\begin{lstlisting}[style=scidocseed]
Inputs: DATASET_LEXICON, ALIAS_MAP, two simulated paper records or IR-mined dataset mentions, and LOCKED_DATASET_SETS.

Check that each selected name plausibly denotes a dataset or scientific resource. Write or polish two short paper-like passages that mention the locked dataset sets without adding another dataset name. Ask for the resources common to both documents and require canonical names. Compute the answer by exact set intersection after ALIAS_MAP normalization. Reject unresolved family/version distinctions or prose that leaks the intersection explicitly.

Answer schema: {"intersection": ["<canonical dataset names>"]}
\end{lstlisting}

\Needspace{5\baselineskip}
\subsubsection{Notation Understanding}

\seedtaskheading{12. Notation extraction.}
\begin{lstlisting}[style=scidocseed]
Inputs: GENERATED_FORMULAS, DEFINITION_SENTENCES, SAVED_SYMBOL_TABLE, and a XeLaTeX page template. Real IR input is not required.

Polish only the unlocked surrounding prose while preserving every formula, symbol, definition, and first-occurrence position. Ask for all symbols explicitly defined in the displayed scope, their definitions, and their first visible locations. Do not include symbols that are merely used. Copy SAVED_SYMBOL_TABLE as the answer and reject a rendering in which a formula or definition is clipped or separated from its scope.

Answer schema: {"notations": [{"symbol": "<LaTeX>", "definition": "<text>", "first_location": "<visible location>"}]}
\end{lstlisting}

\seedtaskheading{13. Symbol ambiguity disambiguation.}
\begin{lstlisting}[style=scidocseed]
Inputs: repeated TARGET_SYMBOL, OCCURRENCE_RECORDS with distinct definitions, formula and context templates, and a one-column, two-column, or spanning page layout.

Naturalize only unlocked context sentences. Preserve each occurrence of TARGET_SYMBOL and its locally defined meaning. Ask the learner to map every labeled occurrence to the correct definition using its paragraph, formula, or section context. Do not imply that the symbol has one global meaning. Copy the occurrence-to-definition mapping from OCCURRENCE_RECORDS and reject contexts that do not uniquely disambiguate every occurrence.

Answer schema: {"symbol": "<LaTeX>", "occurrences": [{"label": "A", "definition": "..."}]}
\end{lstlisting}

\Needspace{5\baselineskip}
\subsubsection{Citation-Role Analysis}

\seedtaskheading{14. A3-lite citation-role classification.}
\begin{lstlisting}[style=scidocseed]
Inputs: real block LaTeX and text, CITATION_KEY, section name, page, block id, local citing context, ROLE_ONTOLOGY, and HEURISTIC_ROLE.

Classify how CITATION_KEY is used in the supplied text, not what the cited paper is generally about. Select exactly one role from Background, MethodBasis, DataResource, Baseline, ResultSupport, Critique, or Extension. Do not consult or invent an external bibliography entry. Return a concise question that includes the observed citing context and the selected role. Reject the sample if several citations share the same grammatical role span, if two roles are equally plausible, or if the classification conflicts with the rule-based prior after review.

Answer schema: {"role": "<one role from ROLE_ONTOLOGY>"}
\end{lstlisting}

The released construction metadata retains source identifiers, locked fields, rendering records, and rejection outcomes. This separates model-assisted linguistic generation from the deterministic or auditable source of each answer.

\clearpage

\section{Experimental Settings and Additional Results}
\label{apx:posttraining}

\subsection{Benchmark Evaluation Setup}

The leaderboard evaluates 13 proprietary and open multimodal models on the same 124 underlying questions under the four settings in Section~\ref{sec:benchmark_contract}, yielding 496 instances per model. The input contains only the question and its source materials; reference answers, task labels, and evaluator metadata are withheld. Overall scores average all instances, while capability-group and setting scores average the corresponding subsets. Failed or unusable responses receive zero rather than being removed from the denominator.

\subsection{Experiment-Specific Training Mixtures}

The reported Qwen3.6 post-training experiments use SciDocDataset. Selection emphasizes scientific-document operations and task-appropriate output contracts, including evidence localization, numerical and symbolic extraction, cross-document integration, and structured or executable outputs.

The SFT targets pair a teacher explanation with the reference final answer. The explanation-generation stage is reference-conditioned: the teacher receives the question and its reference answer and produces an explanatory solution. The corresponding teacher endpoint is recorded as \texttt{qwen3.6-max-preview}. Explanations are generated at the seed--language level and used with the two document views; they are not four independent solutions sampled from the student model. The construction retains existing explanations when regeneration fails or the replacement does not meet the length checks.

For RL, we use the 4,000 question--answer instances from the SFT mixture, without teacher explanations in the verifier references, together with 6,219 additional verifiable QA instances, yielding 10,219 instances. This mixture covers layout relations, tables, charts, notation, and document evidence. The published collection contains 3,924 SFT and 10,143 RL instances; its composition and splits are reported in Table~\ref{tab:bench_stats}.

\subsection{Optimization and Reward}

We perform SFT using LlamaFactory~\citep{zheng2024llamafactory} and GRPO using ms-swift~\citep{zhao2025swift}. SFT applies LoRA to the language component while freezing the vision tower and multimodal projector. GRPO starts from the merged SFT checkpoint and learns a new adapter. Both stages use ZeRO~\citep{rajbhandari2020zero}; the RL loss implementation is Dr.~GRPO~\citep{liu2025drgrpo}. Table~\ref{tab:posttraining_hparams} summarizes the training configuration. The selected checkpoints correspond to 50 SFT optimizer steps and 200 additional GRPO optimizer steps; the latter is not a separate RL-only initialization.

\begin{table}[!tp]
\centering
\caption{\textbf{Post-training configuration.} The RL verifier runs on two additional H200 GPUs alongside six training GPUs.}
\label{tab:posttraining_hparams}

\small
\renewcommand{\arraystretch}{1.16}
\begin{tabularx}{\linewidth}{@{}p{0.40\linewidth}YY@{}}
\toprule
\textbf{Setting} & \textbf{SFT} & \textbf{GRPO} \\
\midrule
\rowcolor{scidocgreen}
\multicolumn{3}{c}{\textit{Model adaptation and optimization}} \\
Initialization & Original Qwen3.6-27B & Merged SFT step 50 \\
Adaptation & LoRA & LoRA \\
LoRA rank / alpha & 32 / 64 & 16 / 32 \\
Learning rate & $1.5\times10^{-5}$ & $5\times10^{-7}$ \\
Learning-rate schedule & Cosine & Cosine \\
Warmup ratio & 0.05 & 0.03 \\
\midrule
\rowcolor{scidocgreen}
\multicolumn{3}{c}{\textit{Compute and memory}} \\
Training GPUs & 8 H200 & 6 H200 \\
Per-device training batch size & 1 & 1 \\
Gradient accumulation steps & 2 & 2 \\
Precision & BF16 & BF16 \\
DeepSpeed stage & ZeRO-3 & ZeRO-2 \\
Gradient checkpointing & Enabled & Enabled \\
\midrule
\rowcolor{scidocgreen}
\multicolumn{3}{c}{\textit{Input, rollout, and reward optimization}} \\
Configured maximum length & 32,768 & 32,768 \\
Maximum rollout completion & -- & 4,096 tokens \\
Maximum image pixels & 589,824 & 262,144 \\
Samples per prompt & -- & 8 \\
Sampling temperature / top-$p$ & -- & 0.8 / 0.95 \\
KL coefficient & -- & 0.01 \\
Loss implementation & Token cross-entropy & \texttt{dr\_grpo} \\
Lower / upper clipping & -- & 0.20 / 0.28 \\
Reward scaling & -- & None \\
\midrule
\rowcolor{scidocgreen}
\multicolumn{3}{c}{\textit{Reproducibility and checkpoint selection}} \\
Random seed & 42 & 42 \\
Saved checkpoint interval & 25 steps & 100 steps \\
\rowcolor{scidocorange}
\textbf{Reported checkpoint} & \textbf{Step 50} & \textbf{Step 200} \\
\bottomrule
\end{tabularx}
\end{table}

\noindent\textbf{Final-answer reward.}
A separate Qwen3.6-27B verifier compares the policy's final answer with the reference under the question's output requirements. It assigns one of three labels: \texttt{CORRECT}, \texttt{PARTIAL}, or \texttt{INCORRECT}, mapped to rewards $1$, $0.2$, and $-1$, respectively. Correctness includes content and material field, format, and ordering requirements; harmless wording differences are permitted. Partial credit requires materially correct progress rather than verbosity alone. The verifier receives the question, reference answer, and candidate final answer, not the policy's internal reasoning. Verifier generation uses temperature zero with thinking disabled. Empty or invalid final answers receive the incorrect reward. The policy itself is allowed to generate a reasoning segment before the final answer.

\subsection{Checkpoint Selection and Evaluation Protocol}

We compare saved checkpoints on a fixed 48-instance diagnostic subset of \ours{} and select SFT step 50 and GRPO step 200. The full benchmark report then evaluates each selected checkpoint and the original model on all 496 instances, including those diagnostic instances. All three full-benchmark runs use the same Transformers~\citep{wolf2020transformers} inference interface, temperature zero, and a maximum of 32,768 newly generated tokens. This budget includes reasoning and the final answer.

The scorer removes the reasoning segment and evaluates only the final response. A truncated reasoning segment without a valid final-answer boundary is not treated as an answer; missing final answers receive zero. Rule-based and execution-based tasks retain their task-specific evaluators. LLM-judged tasks use GPT-5.4-mini with temperature zero and seed 42. All three models use the same final-answer extraction and scoring rules. Raw responses are preserved separately from scored final responses in lossless records.

For the general document benchmarks, we use the standard VLMEvalKit~\citep{duan2024vlmevalkit} implementations on the final answer, with empty answers scored as zero. DocVQA and InfoVQA use their labeled validation splits of 5,349 and 2,801 questions and report average normalized Levenshtein similarity (ANLS). ChartQA uses its 2,500-question test split, including 1,250 human-authored and 1,250 augmented questions. Its relaxed numeric accuracy allows a relative error of 5\% for nonzero numeric references; nonnumeric answers use case-insensitive matching. All scores are on a 0--100 scale, and differences are calculated before rounding.

DocVQA and InfoVQA inference uses a 32,768-token generation cap and temperature zero. The original-model runs use Transformers, SFT uses the completed cached runs, and GRPO uses vLLM~\citep{kwon2023vllm}. Missing final answers are retained and scored as zero, including 19 DocVQA and 51 InfoVQA responses for GRPO.

For ChartQA, all three models use the native VLMEvalKit \texttt{Qwen3VLChat} Transformers backend and built-in benchmark prompt, without a custom prompt. We use Qwen's recommended sampling parameters: \texttt{do\_sample=true}, temperature 1.0, top-$p$ 0.95, top-$k$ 20, repetition penalty 1.0, and a maximum of 32,768 new tokens, including reasoning. Final-answer extraction and relaxed-accuracy scoring are identical across models. No additional training on the DocVQA, InfoVQA, or ChartQA training splits is used in these experiments.

\subsection{Capability, Matched-Setting, and ChartQA Results}

Table~\ref{tab:posttraining} jointly reports the seven capability groups, four matched input settings, and general document scores. SFT improves every matched setting over the original model. The GRPO-adapted model improves both Chinese settings and English interleaved performance over the same baseline. Under the common ChartQA configuration, the original, SFT, and SFT+GRPO models score 74.60, 75.40, and 78.00, respectively. Their gains over the original model are 0.80 and 3.40 points.

\subsection{Qualitative Failures of the Strongest Model}
\label{apx:qualitative_failures}

Table~\ref{tab:qualitative_failures} examines three completed Claude-Opus-5 responses in the English All Images First setting. These are content errors in nonempty responses, rather than API failures. The examples separate errors in task scope, visual interpretation, and completeness; they are illustrative cases, not estimates of their frequency.

\begin{table}[!tp]
\centering
\caption{\textbf{Selected Claude-Opus-5 failures.} IDs refer to Appendix~\ref{app:tasks}; outputs are abbreviated without changing the contrasted values. Scores are on the 0--1 scale.}
\label{tab:qualitative_failures}
\footnotesize
\setlength{\tabcolsep}{4pt}
\renewcommand{\arraystretch}{1.12}
\begin{tabularx}{\linewidth}{@{}p{0.15\linewidth}XX@{}}
\toprule
\textbf{Task} & \textbf{Reference and predicted output} & \textbf{Failure interpretation} \\
\midrule
ID 13: ordered metric extraction (A1); score 0 & Reference: [Chamfer Distance, Precision, Recall, F1-score]. Prediction: [accuracy, completeness, Chamfer Distance (CD), precision, recall, F1-score]. & The model retrieves the target metrics but adds two earlier entries. Since the question restricts the section and order, every aligned position is wrong. This is a scope/ordering failure, not failure to recognize all four metrics. \\
\midrule
ID 59: logarithmic chart readout (B2); score 0.30 & Reference: AQC(exp) range G ($8\!\times\!10^{-4}$ to $5\!\times\!10^{-1}$), QAOA range C. Prediction: AQC(exp) range A ($5\!\times\!10^{-3}$ to $5\!\times\!10^{-1}$), QAOA range C. & The wrong AQC lower endpoint is also asserted in the explanation. Correct QAOA selection earns partial credit; the rubric does not credit an explanation grounded in the incorrect joint range selection. \\
\midrule
ID 71: table consistency audit (C1); score 0.67 & Reference errors: 48.4, 47.8, 32.4. Prediction: 68.0, 32.4, 48.4. The missing 47.8 should be 47.3, the rounded mean of eight entries. & Two errors are found, including a correctly recomputed mean, but another mean is not checked correctly. The additional highlighting complaint about 68.0 does not recover the missing reference error. \\
\bottomrule
\end{tabularx}
\end{table}


\end{document}